\documentclass[10pt,journal,compsoc]{IEEEtran}

\ifCLASSOPTIONcompsoc
  \usepackage[nocompress]{cite}
\else
  \usepackage{cite}
\fi

\ifCLASSOPTIONcompsoc
	\usepackage[caption=false,font=normalsize,labelfon t=sf,textfont=sf]{subfig}
\else
	\usepackage[caption=false,font=footnotesize]{subfig}
\fi

\usepackage{amsmath,amsfonts}
\usepackage{algorithmic}
\usepackage{algorithm}
\usepackage{array}
\usepackage{stfloats}
\usepackage{tabularx}
\usepackage{ulem}
\usepackage{multirow}
\usepackage{booktabs}
\usepackage[pagebackref,breaklinks,colorlinks]{hyperref}
\usepackage{url}
\usepackage{verbatim}
\usepackage{graphicx}
\usepackage{CJKutf8}
\usepackage{paralist}
\usepackage[smartEllipses]{markdown}
\newcommand{\eref}[1]{Eq.~(\ref{#1})}
\newcommand{\fref}[1]{Figure~\ref{#1}}
\newcommand{\sref}[1]{Section~\ref{#1}}
\newcommand{\tref}[1]{Table~\ref{#1}}
\newcommand{\PreserveBackslash}[1]{\let\temp=\\#1\let\\=\temp}
\newcolumntype{C}[1]{>{\PreserveBackslash\centering}p{#1}}

\begin{document}

\title{StructVPR++: Distill Structural and Semantic Knowledge with Weighting Samples for Visual Place Recognition}

\author{Yanqing~Shen,
        Sanping~Zhou,~\IEEEmembership{Member,~IEEE,}
        Jingwen~Fu,
        Ruotong~Wang,
        Shitao~Chen,~\IEEEmembership{Member,~IEEE,}
~and~Nanning~Zheng,~\IEEEmembership{Fellow,~IEEE}
\IEEEcompsocitemizethanks{\IEEEcompsocthanksitem Yanqing Shen, Sanping Zhou, Jingwen Fu, Ruotong Wang, Shitao Chen, and Nanning Zheng are with the National Key Laboratory of Human-Machine Hybrid Augmented Intelligence, National Engineering Research Center for Visual Information and Applications, and Institute of Artificial Intelligence and Robotics, Xi'an Jiaotong University, Xi’an, Shaanxi 710049, China. E-mail: 
\{qing1159364090@stu., spzhou@mail.,  fu1371252069@stu., wrt072@stu., chenshitao@, nnzheng@mail.\}xjtu.edu.cn
(Corresponding author: Nanning Zheng.)
}
}

\markboth{Journal of \LaTeX\ Class Files,~VOL. XX, NO. XX, XX XX}%
{Shen \MakeLowercase{\textit{et al.}}: XXXX}

\begin{CJK*}{UTF8}{gbsn}

\newpage

\IEEEtitleabstractindextext{
\begin{abstract}

Visual place recognition is a challenging task for autonomous driving and robotics, which is usually considered as an image retrieval problem.
A commonly used two-stage strategy involves global retrieval followed by re-ranking using patch-level descriptors.
Most deep learning-based methods in an end-to-end manner cannot extract global features with sufficient semantic information from RGB images.
In contrast, re-ranking can utilize more explicit structural and semantic information in one-to-one matching process, but it is time-consuming.
To bridge the gap between global retrieval and re-ranking and achieve a good trade-off between accuracy and efficiency, we propose StructVPR++, a framework that embeds structural and semantic knowledge into RGB global representations via segmentation-guided distillation.
{Our key innovation lies in decoupling label-specific features from global descriptors, enabling explicit semantic alignment between image pairs without requiring segmentation during deployment.}
{Furthermore, we introduce a sample-wise weighted distillation strategy that prioritizes reliable training pairs while suppressing noisy ones.}
{Experiments on four benchmarks demonstrate that StructVPR++ surpasses state-of-the-art global methods by 5-23\% in Recall@1 and even outperforms many two-stage approaches, achieving real-time efficiency with a single RGB input.}

\end{abstract}

}

\maketitle

\IEEEdisplaynontitleabstractindextext

\IEEEraisesectionheading{\section{Introduction}}
\label{sec:introduction}



\IEEEPARstart{V}{isual} place recognition (VPR) is a pivotal task in autonomous driving and robotics, typically framed as an image retrieval problem~\cite{lowry2015visual,zhang2021visual,masone2021survey}.
Given a query RGB image, VPR aims to determine whether {it originates from a previously visited location by matching it against a database of reference images.}
{However, this task faces challenges due to extreme environmental variations, including long-term changes and transient disturbances such as seasonal shifts, illumination differences, vegetation growth, and dynamic occlusions.}
{To address these challenges, developing discriminative and robust features becomes critical for distinguishing locations.}

\begin{figure}[t]
  \centering
\subfloat[]{
  \label{fig:seg}
  \includegraphics[width=0.45\columnwidth]{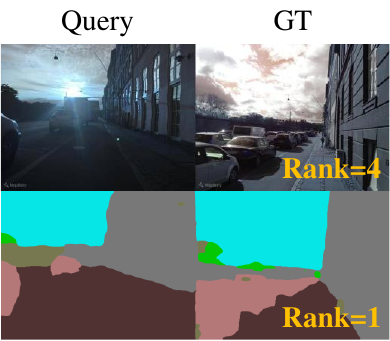}}
\subfloat[]{
  \label{fig:rgb}
  \includegraphics[width=0.45\columnwidth]{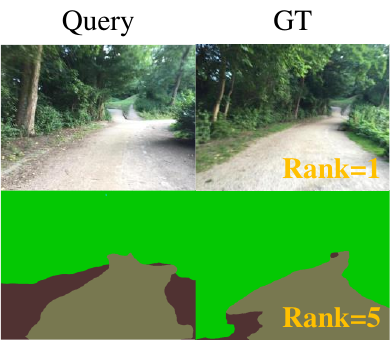}}
   \caption{\textbf{Examples of query images and ground truths.} The marked number represents the recall performance of two pre-trained branches on ground truths. (a) shows the scene with illumination variation and seasonal changes, where segmentation images are more recognizable. (b) shows the scene with changing perspectives, where RGB images are more recognizable.}
   \label{fig:show}
\end{figure}

\begin{figure}
    \centering
    \includegraphics[width=0.98\linewidth]{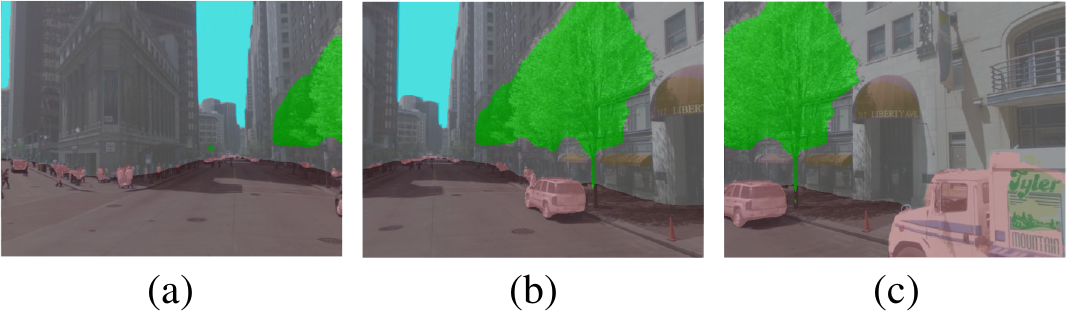}
    \caption{{Challenging cases with large viewpoint variations from the same place. Existing methods often fail to align semantic meanings between image pairs, but it's difficult to only utilize structural information to complete recognition when lack of overlap.}}
    \label{fig:misalignment}
\end{figure}

{Existing approaches often adopt a two-stage pipeline: global feature-based candidate retrieval followed by local descriptor re-ranking~\cite{transvpr,r2former}.}
Although re-ranking is time- and resource-consuming, it significantly improves performance by utilizing rich structural information {such as shape, edge~\cite{wasabi}, spatial layout, and category\cite{xview}, which is more robust to VPR than global appearance information.}
Recent studies~\cite{dasgil,bao2023hierarchical}, including our prior work~\cite{structvpr}, have illustrated that segmentation (SEG) images can provide effective structural cues for VPR.
{Notably, as illustrated in \fref{fig:show}, SEG and RGB image modalities complement each other: SEG excels in illumination and seasonal variations, whereas RGB handles viewpoint changes.}
{Nevertheless, structural information alone remains inadequate for severe viewpoint shifts or occlusions—common challenges in real-world applications, as shown in \fref{fig:misalignment}.}

{In this paper, we explore semantic misalignment, where similar features between image pairs originate from semantically distinct regions, leading to erroneous correlations in global image representations.}
{To mitigate this, we propose integrating segmentation-derived semantic consistency into the network architecture.} Specifically, segmentation images provide explicit label-level supervision, enabling the model to enforce semantic alignment during feature learning.

{Motivated by these insights, we present StructVPR++\footnote{\href{https://github.com/syqlyx/StructVPR}{https://github.com/syqlyx/StructVPR}}, a distillation framework that transfers structural and semantic knowledge from segmentation images to RGB representations through additional weighted label features.}
{As shown in \fref{fig:pipeline}, the framework operates in two phases: (1) training separate branches for segmentation and RGB feature extraction, called \textit{seg-branch} and \textit{rgb-branch}, followed by (2) selective knowledge distillation from SEG to RGB.}
StructVPR++ consists of four components: segmentation label map encoding, feature extraction, group partition, and weighted knowledge distillation.

In the first training stage, {segmentation images are converted into standard CNN inputs.}
{Before that, we group semantically similar categories and assign importance weights to key scene elements. This reduces label noise and helps the model focus on stable structural patterns.}
Next, {the seg-branch and rgb-branch are independently trained with triplet loss.}

{Addressing sample heterogeneity is crucial for effective distillation, as some samples provide valuable knowledge while others introduce noise.}
{To tackle this, we propose a novel group partitioning strategy that leverages frozen seg-branch and rgb-branch from Stage I to classify samples into distinct groups.}
{Experimental analysis reveals that these groups impact distillation efficacy differently, enabling targeted knowledge transfer.}

In the second training stage, {the frozen seg-branch serves as the teacher network, transferring knowledge to the label-aware RGB student model.}
A sample-weighting function is designed to dynamically prioritize beneficial samples while suppressing noisy ones during distillation.
Then the {label-aware RGB model} is jointly optimized using distillation loss and standard VPR loss.
Crucially, during deployment, only the label-aware RGB model is used, avoiding segmentation overhead while ensuring efficient place recognition without sacrificing robustness to environmental changes.

In summary, the key contributions of this paper include
\begin{enumerate}[1.]
    \item We explore the potential of extracting structural and semantic cues for place recognition from segmentation images. To resolve semantic misalignment, we decouple label-specific features from global representations.
    \item A weighted knowledge distillation pipeline is proposed, {which enhances segmentation-guided knowledge into RGB models, while eliminating the need for segmentation during deployment.}
    \item {A new group partition strategy, combined with a sample-weighting function, is introduced to dynamically prioritize beneficial samples during distillation.}
    \item {Extensive experiments across four datasets show consistent improvements, with absolute Recall@1 gains of 5-23\% over the best global baseline.}
\end{enumerate}

This paper extends our previous conference paper~\cite{structvpr}, and the new major contributions include:

\begin{enumerate}[1.]
    \item {We explore semantic misalignment and analyze structural matching failures under extreme viewpoint changes and occlusions.}
    \item We dig into the usage of segmentation images and extend the knowledge extraction module by separating label features from global features, summarized as in contribution 1. It explicitly incorporates semantic information into global representations.
    \item More technical details and design ideas about the framework along with discussions on its strengths and limitations, are provided. Additionally, more qualitative results are given, including visualizations of semantic misalignment and the effect of weighted label features.
    \item Last but not least, more extensive experiments are carried out to compare our method with its ablated variants and evaluate its effectiveness.
\end{enumerate}

The rest of the paper is organized as follows. Section 2 discusses related work in visual place recognition and knowledge distillation. Subsequently, we present the technical details of the proposed method in Section 3. Experimental results and discussions are presented in Section 4. Finally, we conclude this paper in Section 5.
\section{Related work}
We briefly review related work in visual place recognition, attentions and semantics for place recognition, and knowledge distillation.

\subsection{Visual Place Recognition}

Most of the research on VPR has focused on constructing better image representations to perform retrieval.
The common way to represent a single RGB image is to use global descriptors or local descriptors.
In early VPR systems\cite{mur2017orb, seqnet}, the hand-crafted local features such as SIFT \cite{lowe1999object}, SURF \cite{bay2006surf} and ORB \cite{rublee2011orb} have been widely used to detect and describe keypoints, while they cannot handle severe appearance changes.
Then more recent CNN-based local features have achieved superior performances\cite{noh2017large, khaliq2019holistic, teichmann2019detect,camara2020visual}.
Moreover, SuperPoint \cite{detone2018superpoint} and D2Net \cite{dusmanu2019d2} use shared network to preform both detection and description.

Global descriptors can provide a compact representation by aggregating local descriptors or directly processing the whole image. Global descriptors can be generated by directly extracting\cite{zhou1,chen2017deep,gordo2017end,radenovic2018fine,revaud2019learning} or aggregating local descriptors. 
For aggregation, traditional methods, such as Bag of Words (BoW)\cite{sivic2003video}, Fisher Kernel\cite{perronnin2007fisher,perronnin2010large, jegou2011aggregating}, and Vector of Locally Aggregated Descriptors (VLAD)\cite{jegou2010aggregating,arandjelovic2013all}, have been used to aggregate hand-crafted local features through a visual codebook and then have been incorporated into CNN-based architectures \cite{netvlad, mohedano2016bags}.
Other CNN-based works mainly focused on designing specific pooling layers on top of convolutional trunks \cite{pool1}.

To achieve a good compromise between accuracy and efficiency, a widely used architecture is to rank the database by global features, and then re-rank the top candidates\cite{sta,sarlin2020superglue,tcl,hausler2021patch}.
Patch-NetVLAD \cite{hausler2021patch} uses NetVLAD as their backbone for global retrieval and applies an integral feature space to derive patch descriptors from global features.
Other local matching methods from related tasks, e.g. SuperGlue \cite{sarlin2020superglue} and DELG \cite{delg}, are also evaluated for VPR.
Recently, unified networks have been developed to jointly extract global and local descriptors\cite{taira2018inloc,sarlin2019coarse,simeoni2019local, transvpr,r2former}.
R2Former\cite{r2former} integrates retrieval and reranking into a unified framework with pure transformers, achieving SOTA performance.
However, additional re-ranking is time-consuming, and our work mainly focus on enhancing knowledge from segmentation images into RGB global features.

\subsection{Semantics for Place Recognition}
In the early stage of research, some work~\cite{add} innovatively take the lead in considering local image semantic contents, thereby eliminating many false matches.
Recently, many methods try to introduce semantics into RGB features by using attention mechanism or additional supervision. 
TransVPR introduces the attention mechanism to guide models to focus on invariant regions and extract robust representations.
{DASGIL uses multi-task architecture with a single shared encoder to create global representation, and uses domain adaptation to align models on synthetic and real-world datasets.
Based on DASGIL, a work focuses on filtering semantic information via an attention mechanism.}
SFD2 \cite{sfd2} focuses on effectively leveraging high-level semantics for low-level feature extraction and creatively uses explicit semantics as supervision to guide the detection and description behaviors.
OrienterNet \cite{orienternet} firstly proposes the deep neural network capable of localizing an image with sub-meter accuracy using the same 2D semantic maps that humans use. It is closer to practical applications.
StructVPR\cite{structvpr} uses a network to directly extract global structural knowledge from segmentation images, and enhance them in RGB representations through sample-based weighted KD.

In this paper, we believe that segmentation images are worthy of further study in VPR to achieve semantic alignment by providing label-level supervision, which is deeper than structural information.

\subsection{Knowledge Distillation} 
It is an effective way to enrich models with knowledge distillation (KD)\cite{gou2021knowledge}. 
It extracts specific knowledge from a stronger model (i.e., ``teacher'') and transfers it to a weaker model (i.e., ``student'') through additional training signals.
There has been a large body of work on transferring knowledge with the same modality, such as model compression\cite{bucilu2006model,chen2017learning,hinton2015distilling} and domain adaptation\cite{li2017large,asami2017domain}. 
However, the data or labels for some modalities might not be available during training or testing, so it is essential to distill knowledge between different modalities \cite{ren2021learning,zhao2020knowledge}.
\cite{garcia2018modality,hoffman2016learning} generate a hallucination network to model depth information and enforce it for RGB descriptors learning. In this way, the student learns to simulate a virtual depth that improves the inference performance.
In this work, we construct a weighted knowledge distillation architecture to distill and enhance high-quality structural knowledge into RGB features.

Nevertheless, these works enhance the target model by transferring knowledge on all training samples from the teacher model, rarely discussing the difference in knowledge among samples \cite{liang2021reinforced,ge2018low}.
Wang \textit{et al.} \cite{selective} proposes to select suitable samples for distillation through analyzing the teacher network.
Differently, our solution considers both pre-trained teacher and student networks in sample partition and weights the distillation loss for samples.



\section{Proposed Method}

\begin{figure*}
    \centering
    \includegraphics[width=\linewidth]{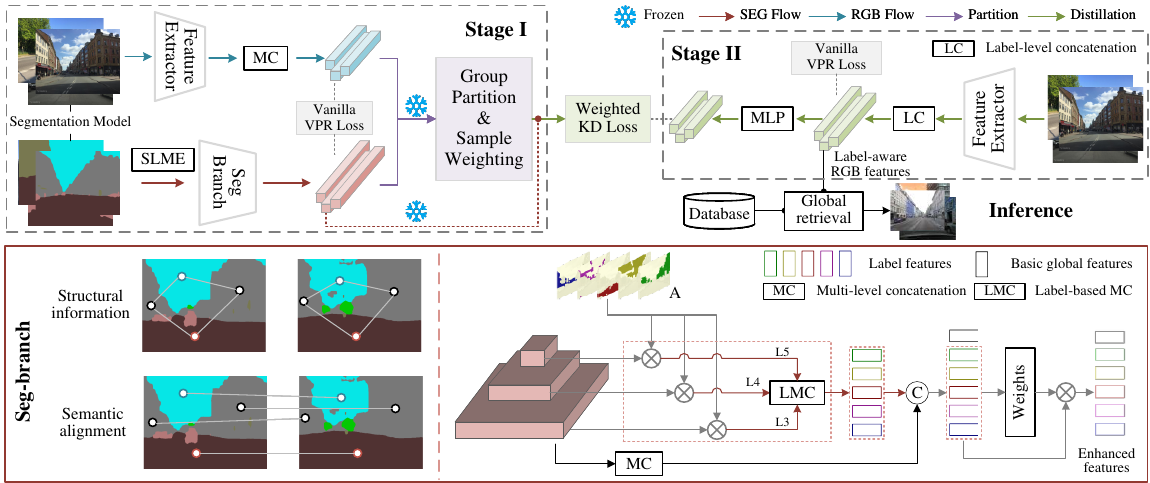}
    \caption{\textbf{Illustration of the proposed pipeline.} We first train two branches with VPR supervision to extract structural and semantic knowledge, respectively. Next, offline group partition is performed using the frozen branches, and weights are assigned to the samples. Then weighted knowledge distillation and VPR supervision are performed in Stage II to train the RGB model using combined loss, and the label-aware RGB features can be obtained. During inference phase, StructVPR++ only uses the trained model in Stage II and does not perform segmentation. More importantly, it can maximize the efficiency of distilling high-quality knowledge. SLME represents the pre-coding process of segmentation images into standard CNN input format. MLP, LC, MC, and LMC are network structure modules used to obtain corresponding features. In seg-branch, \textbf{A} refers to the multi-level down-sampled masks, and \textbf{Weights} means the shared weighting module for separate label features. The schematic diagram on the lower left briefly illustrates the macro differences between the structural and semantic information.}
    \label{fig:pipeline}
\end{figure*}

{Motivated by the observation that local feature re-ranking injects explicit structural priors to enhance performance, we incorporate segmentation images to enhance the rich and explicit structural knowledge int RGB representations.
Our method bridges the performance gap with two-stage approaches while maintaining single-stage efficiency.}
{Additionally, we employ label-level supervision to improve semantic alignment across the entire framework.}
The key idea of StructVPR++ consists of two parts: selectively distill high-quality knowledge into a single-input model through sample weighting, and separate label features from global features to enhance semantic alignment.
StructVPR++ is trained in two stages: structural knowledge extraction and group partition (Stage I), and weighted knowledge distillation (Stage II). \fref{fig:pipeline} explains the training process of StructVPR++ and how StructVPR++ works in the inference phase.

\subsection{Overview}
\label{subsec:over}
Given an input RGB image $I^{R}$, its associated segmentation image $I^{S}$ can be extracted by an open-source semantic segmentation model and converted into a fixed-size label map via \textit{segmentation label map encoding} (SLME) module.

To better achieve semantic alignment, seg-branch is designed to explicitly extract label features, and then the seg-branch and rgb-branch are trained with VPR loss in Stage I.
Based on the two trained branches in Stage I and the designed evaluation metric, the whole training set can be partitioned offline, and different groups represent samples with different performances of seg-branch and rgb-branch. 
Given the domain knowledge in VPR, the concept of ``sample'' is defined as \textit{a sample pair of a query and a positive}.
There are two specific aspects of domain knowledge:
First, VPR datasets are often organized into \textit{query} and \textit{database}, in which each query image $q$ has a set of positive samples $\{p^{q}\}$ and a set of negative samples $\{n^{q}\}$.
Second, the knowledge to be enhanced is feature invariance and robustness under changes between sample pairs in a scene.

In Stage II, the label-aware RGB model, different from rgb-branch, {is designed to incorporate label features.}
{A weighting function is introduced to facilitate the distillation} of enhanced SEG features into the label-aware RGB model.
We perform weighted knowledge distillation and standard VPR supervision in this stage.
{During inference, we utilize label-aware global RGB features for improved semantic alignment, eliminating the need for online segmentation.}

\subsection{Segmentation Label Map Encoding}
\label{subsec:encoding_module}
It is necessary to convert segmentation images to a standard tensor format before inputting them into the seg-branch.
As shown in \fref{fig:pipeline}, the segmentation label map encoding (SLME) function encodes segmentation information into a weighted one-hot representation, including three steps: formatting, incorporating, and weighting.

Firstly, like LabelEnc\cite{hao2020labelenc}, we use a $C \times H \times W$ tensor to represent segmentation images, where $H \times W$ equals the RGB image size and $C$ is the number of semantic classes.
Regions of the c-th class are filled with positive values in the c-th channel and 0 in other channels.

\begin{figure}[t]
    \centering
    \includegraphics[width=0.9\linewidth]{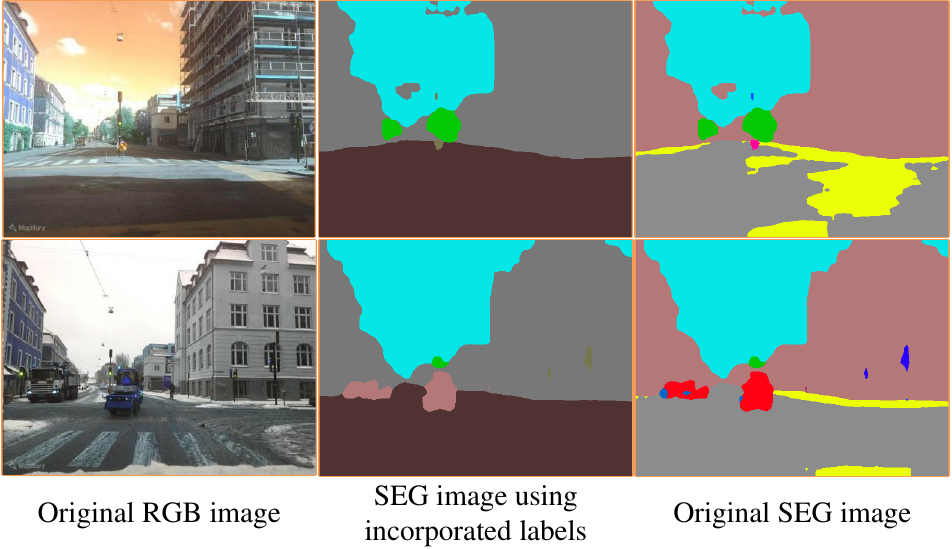}
    \caption{\textbf{Visualization of incorporated semantic classes.} Shown from left to right are RGB images, segmentation images with incorporated labels, and original segmentation images. {It can be seen that the label space after incorporating is cleaner for VPR.}}
    \label{fig:cluster}
\end{figure}

{Secondly, to avoid over-segmentation noise~\cite{paolicelli2022learning} and reduce computation overhead, SLME adopts coarse-grained labels (e.g., merging ``cars'' and ``bicycles'' into ``dynamic objects'').
Specifically, we incorporate the semantic labels from the original 150 to 6 based on human experience.}
In particular, ``dynamic'' category is completely abandoned and final $C=5$ in this work, which means that the vectors corresponding to pixels of the ``dynamic'' label are encoded as all 0 vectors.
This step reduces computational costs and lets models focus on useful parts.
More details about these categories are provided in \sref{sec:details} \textit{Implementation Details}.

{Finally, SLME incorporates domain knowledge by assigning distinct weights to semantic categories. 
This design stems from the observation that distinct semantic classes (e.g., buildings vs. vegetation) contribute differentially to place recognition, allowing the model to amplify stable structural cues while suppressing transient elements.}
In~\sref{sec:ablation}, experimental results corroborate our idea.

\subsection{Structural Knowledge Extraction}
\label{subsec:branch}

In Stage I, we train seg-branch and rgb-branch to extract structural and semantic knowledge into features, which will be used for subsequent group partition.

For rgb-branch, it is difficult to converge and learn effective information by explicitly decoupling label features from global features only through image-level VPR supervision. So, only {basic global features} are extracted in Stage I.

In order to solve the problem of semantic misalignment in the global features extracted by seg-branch, we divide the {enhanced global features} into two parts, {basic global feature}s and {weighted label features}, thereby explicitly introducing label features into enhanced global features.

\subsubsection{Rgb-branch}
\label{sec:rgb}

\textbf{Backbone.} We use MobileNetV2\cite{MobileNetV2} as our lightweight extractor. We remove the global average pooling layer and fully connected (FC) layer to obtain global features. According to the resolution of the feature maps, the backbone can be divided into 5 stages. 
The output feature maps are denoted as $\mathcal{F}^{R}$: \{$\mathcal{F}^{R}_{1}$, $\mathcal{F}^{R}_{2}, \mathcal{F}^{R}_{3}, \mathcal{F}^{R}_{4}, \mathcal{F}^{R}_{5}$\}, and the global features are denoted as 
\begin{equation}
    \mathbf{x}^R = \mathrm{MC}(\mathcal{F}^{R}),
\end{equation}
which are trained with VPR loss.

\begin{figure}[t]
    \centering
    \includegraphics[width=0.95\linewidth]{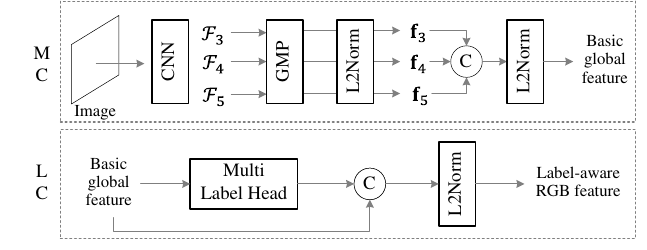}
    \caption{\textbf{Flowchart of feature computation.} Visualization of multi-level concatenation layer and label-level concatenation layer.}
    \label{fig:mc}
\end{figure}

\textbf{Multi-level Concatenation (MC).}
In \fref{fig:mc}, we first apply a global max pooling (GMP) layer to each feature map, $\mathcal{F}_{i} \in \mathbb{R}^{H_i \times W_i \times C_i}$, to compute single-level features, like 
\begin{equation}
    \mathbf{f}_{i} = \mathrm{L2Norm}(\mathrm{GMP}(\mathcal{F}_{i})) \in \mathbb{R}^{ C_i \times 1},
\end{equation}
and concatenate the features $\mathbf{f}_{i} $ from the last 3 layers:
\begin{equation}
\begin{aligned}
    \mathbf{x} &= \mathrm{L2Norm}(\mathrm{Concat}([\mathbf{f}_{3},\mathbf{f}_{4},\mathbf{f}_{5}]))
    = \mathrm{MC}(\mathcal{F}), \\
\end{aligned}
\end{equation}
where $\mathbf{x}$ is the unified representation of basic global features.

\textbf{Triplet VPR Loss.} 
In two training stages, we adopt triplet margin loss\cite{schroff2015facenet} on $(q, p, n)$ as supervision:
\begin{equation}
    \mathcal{L}_{vpr}(\mathbf{x})\!=\!\mathrm{max}(d(\mathbf{x}_q,\mathbf{x}_p)-d(\mathbf{x}_q,\mathbf{x}_n)+m,0), \label{eq:triplet_loss}
\end{equation}
where $\mathbf{x}_q$, $\mathbf{x}_p$, and $\mathbf{x}_n$ refer to global features of query, positive and negative samples. $d(\cdot)$ computes the $L2$ distance of two feature vectors, and margin $m$ is a constant parameter.

\subsubsection{Seg-branch}
\label{sec:seg}

\textbf{Backbone and Basic Global Features.} Considering that segmentation label maps are more straightforward and have a higher semantic level than the associated RGB images, we adopt the depth stream in MobileSal\cite{mobilesal} as the feature extractor in seg-branch. It has five stages with same strides and is not as large a capacity as MobileNetV2, denoted MobileNet-L in our paper. 
The output feature maps of five stages are denoted as $\mathcal{F}^{S}$: \{$\mathcal{F}^{S}_{1}, \mathcal{F}^{S}_{2}, \mathcal{F}^{S}_{3}, \mathcal{F}^{S}_{4}, \mathcal{F}^{S}_{5}$\}, and the basic global feature is represented as
\begin{equation}
    \mathbf{x}^S = \mathrm{MC}(\mathcal{F}^{S}) \in \mathbb{R}^{D_s \times 1},
\end{equation}
where $D_s$ is the latent embedding dimension.



\textbf{Weighted Label Features.} 
Since the global features obtained by previous deep learning-based methods are at the image-level, there is no guarantee that the calculated values of the same channel of an image pair come from the same label, which is called semantic misalignment in this paper.
The main extension of our method compared with its conference version \cite{structvpr} lies in the label feature generation in seg-branch and subsequent corresponding distillation processing to achieve better semantic alignment during global retrieval.

The multi-level down-sampled segmentation images can be used as a set of multi-level masks $\mathbf{A}_{i,j}, j\in{[1,..,C]}$, and the size of $\mathbf{A}_{i,j}$ is same as feature maps $\mathcal{F}^{S}_{i}$.
We first superimpose the label masks on the feature maps.

\textbf{Label-level MC (LMC)} means that, for certain label $j$,
\begin{equation}
        \mathcal{F}^{S}_{j} = \{ \mathcal{F}^{S}_{i,j}\} = \{ \mathbf{A}_{i,j} * \mathcal{F}^{S}_{i} \}, i\in{[3,4,5]}.
\end{equation}
Followed by MC layer in \sref{sec:rgb}, vanilla label features can be obtained by:
\begin{equation}
        \mathbf{l}^{S}_{j} = \mathrm{MC}(\mathcal{F}^{S}_{j}) \in \mathbb{R}^{D_s \times 1}.
\end{equation}

In addition, we find that the similarity of the global features of the two images is almost consistent with the similarity of the label features, which means that there is no additional information will be extracted after generating vanilla label features in the training.

Based on the above analysis, we propose \textbf{weighted label feature} in this paper.
Moreover, we believe that this design can help the model learn the different role of labels in specific images.
{As shown in \fref{fig:pipeline}}, each label feature passes through the shared MLP layer to obtain the corresponding weight, and then the set of weights passes through the softmax layer:
\begin{equation}
    w_{j} = \mathrm{SoftMax}(\mathrm{MLP}(\mathbf{l}^{S}_{j})).
\end{equation}

\textbf{Enhanced Global Features.} 
To add constraints between the corresponding label features of the image pair, we jointly train basic global features and label features.
Therefore, enhanced global features for seg-branch can be obtained as follows:
\begin{equation}
\label{eq:seg}
    \mathbf{g}^{S} = \mathrm{L2Norm}(\mathrm{Concat}([\mathbf{x}^{S}, w_{j}*\mathbf{l}^{S}_{j}])) \in \mathbb{R}^{(C+1)D_s \times 1},
\end{equation}
where $C$ means the number of semantic classes. As a whole, $\mathbf{g}^{S}$ will be trained with VPR loss. In this way, label features are able to extract common attributes within the certain label, which are consistent for the same scene and unique in different scenes. In the meantime, basic global features can still extract structural knowledge of scenarios for recognizing.

\textbf{Training strategy.}
Firstly, the backbone is initialized by supervising basic global features with triplet VPR loss.
After initialization, the whole seg-branch (enhanced global features) can be further finetuned in an end-to-end fashion on VPR datasets.



\subsection{Group Partition on Training Set}
\label{subsec:group}

As mentioned above, not all samples contain high-quality and helpful knowledge for distillation, and even some will hurt the performance.
This section addresses this problem simply yet effectively by selecting suitable samples to distill the expected knowledge. 

Previous sample-based selective distillation works\cite{selective,liang2021reinforced,ge2018low} rarely consider the student network with specific prior knowledge, resulting in a lack of teacher-student interaction in cross-modal cases.
To avoid this, we let the two trained branches in \sref{subsec:branch} participate in seeking a more accurate partition.

\textbf{Partition Strategy.}
For convenience, ``samples'' mentioned below refer to \textit{sample pairs} (see \sref{subsec:over}).
In this section, we first did some preliminary empirical research, and we found that trained seg-branch and rgb-branch have low VPR loss on training sets, but their Recall@$N$ performance cannot achieve 100$\%$ on the whole training sets.
Therefore, the performance of the two branches on a sample, $(q,p)$, can be compared by following metric: \textbf{\textit{the ranking of the positive $p$ in the recall list of the query $q$}}.

Based on the frozen seg-branch and rgb-branch trained in stage I, retrieval process can be performed on the entire training set. The entire training set can then be partitioned offline using the designed evaluation metric.
For convenience, recall rankings of samples of seg-branch and rgb-branch are denoted as $x$ and $y$, respectively.
As shown in \fref{fig:pipeline}, specific definitions for partitioning training sets are as follows:
$$
\begin{aligned}
{\mathcal{D}_{1}} &=\{(q,p)|x \leq {N_t},y>{N_t}\}, \\
{\mathcal{D}_{2}} &=\{(q,p)|x \leq y \leq {N_t}\}, \\
{\mathcal{D}_{3}} &=\{(q,p)|y<x \leq {N_t}\}, \\
{\mathcal{D}_{4}} &=\{(q,p)|x>{N_t}\},
\end{aligned}
$$
where ${\mathcal{D}_{1}}$ represents the set of samples with $x$ less than or equal to ${N_t}$ and $y$ greater than ${N_t}$, and so on. ${N_t}$ is a constant hyper-parameter.
Intuitively, ${\mathcal{D}_{1}}$ is the most essential and helpful group for knowledge distillation.

\subsection{Weighted Knowledge Distillation}
\label{subsec:kd}


\subsubsection{Label-aware RGB Model}
\label{sec:model}
In Stage II, the backbone structure of RGB feature extractor is also MobileNetV2, and the basic global feature is same as rgb-branch in \sref{sec:rgb}, denoted as $\mathbf{x}^{R} \in \mathbb{R}^{D_r \times 1}$.
In order to better distill the seg-branch features into the RGB model in Stage II, the RGB model in Stage II should also have the ability to generate label features, thereby allowing the model to learn coarse-grained recovery of segmentation through feature-level distillation.

As shown in \fref{fig:pipeline}, label features are generated by $\mathbf{x}^{R}$ through a set of MLP layers as heads:
\begin{equation}
    \mathbf{l}^{R}_{j} = \mathrm{MLP}_{j}(\mathbf{x}^{R}), j \in [1,..,C],
\end{equation}
where $\mathbf{l}^{R}_{j}$ will be mapped to the domain of $\mathbf{l}^{S}_{j}$ for alignment during the Stage II.

Finally, label-aware global RGB features can be calculated as follows:
\begin{footnotesize}
\begin{equation}
    \mathbf{g}^{R} = \mathrm{LC}(\mathbf{x}^{R}) = \mathrm{L2Norm}(\mathrm{Concat}([\mathbf{x}^{R}, \mathbf{l}^{R}_{j}])) \in \mathbb{R}^{(C+1)D_r \times 1},
\end{equation}
\end{footnotesize}
where $\mathrm{LC}$ means \textbf{label-level concatenation.}

\subsubsection{Sample-based Weighted Distillation}
In order to accurately reflect the impacts of samples, here we weigh the distillation loss based on group partition in \sref{subsec:group}. This way, helpful samples are emphasized, and harmful samples are neglected.

As a preliminary research, we first attempted to apply different constant weights to different groups.
This preliminary experiment is mainly for providing a numerical reference for the design of our weight function.

\textbf{Weighting Function.}
Instead of using several constant weights for different groups or learning-based methods, we define a data-free function, $\varphi$, to refine the weights on samples from two empirical perspectives.
One is the knowledge levels of the teacher on each sample; the higher the knowledge level, the greater the weight. 
Another is the knowledge gap between the teacher and the student; the greater the gap, the greater the weight.

\begin{figure}[b]
    \centering
    \includegraphics[width=0.7\linewidth]{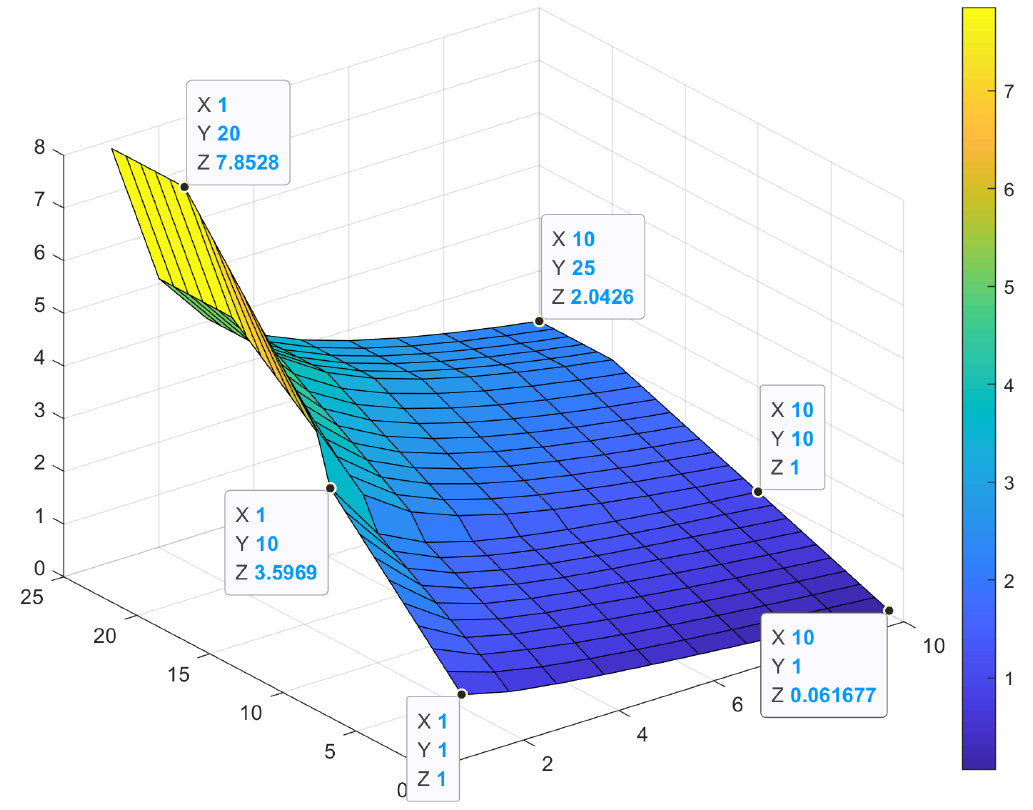}
    \caption{Visualization of the weight function. Some weights of  demarcation points are marked. $X$ and $Y$ mean the recall rankings of samples of seg-branch and rgb-branch.}
    \label{sfig:weight}
\end{figure}

Based on the reference constant weights, the design of function $\varphi$ mainly includes the following considerations:
\begin{itemize}
    \item $\varphi$ cannot be negative;
    \item The non-zero part of $\varphi$ should be proportional to $y-x$ and inversely proportional to $x$;
    \item The value of $\varphi$ cannot be too large;
    \item The partial derivatives should be different for 3 groups.
\end{itemize}


The function has many alternative options that meet the requirements. 
Throughout the design process, we did not perform rigorous tuning of the parameters, but simply chose representative design to demonstrate our insight and motivation.
Finally we analyse two common forms: the prototype of the function can be denoted as $\frac{f(y-x)}{g(x)}$, where $f(\cdot)$ and $g(\cdot)$ are monotonically increasing functions. 
Considering $x$ should play a more important role in weights than $y-x$, we choose linear for $f(\cdot)$ and natural logarithm for $g(\cdot)$ as an improved option (see \fref{sfig:weight}).

Finally, the weight function $\varphi$ is defined as:
\begin{equation}\vspace{-0.05cm}
\label{eq:rule}
\varphi(q,p)=\\
\begin{cases}
1+\frac{\min \left(N_{m},y\right)-x}{4 \cdot \ln \left(1+x\right)}, &(q,p) \in {\mathcal{D}_{1}}\\
1+\frac{y-x}{5 \cdot \ln \left(1+x\right)}, &(q,p) \in {\mathcal{D}_{2}}\\
1+\frac{y-x}{4 \cdot \ln \left(1+x\right)}, &(q,p) \in {\mathcal{D}_{3}}\\
0, &(q,p) \in {\mathcal{D}_{4}}
\end{cases}, \vspace{-0.05cm}
\end{equation}
where {$N_m$ is a hyper-parameter, mainly used to limit the maximum weight}. Here $\varphi$ is proportional to the performance of the seg-branch on $(q,p)$.
This function makes each sample determine its learning degree. 

In \tref{tbl:function}, we show more ablation experiments by replacing $\frac{y-x}{ln(x+1)}$ with $\frac{y-x}{x}$ in \eref{eq:rule},
and it can be seen that our function performs better than the prototype function, showing the advantages of \eref{eq:rule}.

\textbf{Feature-based Distillation.}
Because $\mathbf{g}^{R}$ and $\mathbf{g}^{S}$ for an image pair $(I^{R}, I^{S})$ are not in the same domain and same dimension, we embed $\mathbf{g}^{R}$ via an additional shared linear layer in Stage II, called transformation function \textit{T}:

\begin{footnotesize}
\begin{equation}
T(\mathbf{g}^{R}) = \mathrm{Concat}([\mathrm{MLP}(\mathrm{L2Norm}(\mathbf{x}^R)), \mathrm{MLP}(\mathrm{L2Norm}(\mathbf{l}_{j}^R))]),
\end{equation}
\end{footnotesize}
where $\mathrm{MLP}$ is used to align independent label features from RGB feature space to SEG feature space respectively.

We adopt loss\cite{gupta2016cross} in our distillation process:
\begin{equation}\vspace{-0.05cm}
\label{eq:skt}
\mathcal{L}_{kd}(I)=\varphi(q,p) \cdot\left\|\mathbf{g}^{S}_{I}-T\left(\mathbf{g}^{R}_{I}\right)\right\|_{2}^{2},  I \in \{q,p,n\}, \vspace{-0.05cm}
\end{equation}
where $n$ is the negative sample corresponding to $(q,p)$ in the second-stage training. Note that $\mathbf{g}^{S}_{I}$ corresponds to the trained seg-branch in Stage I.

\begin{table}[t]
  \caption{Comparisons of Different Weighting Functions on MSLS Datasets}
  \label{tbl:function}
  \centering
  \scalebox{1.0}{\begin{tabular}{c|c|c}
  \toprule
  &MSLS val & MSLS challenge \\ \hline
  $\frac{y-x}{x}$ &  82.1/89.8/92.7 &64.2/79.9/83.6\\ \hline
  $\frac{y-x}{ \ln (x+1)}$ &84.3/91.5/93.1&65.7/81.4/85.3\\
  \bottomrule                  
  \end{tabular}}
\end{table}

Hence, the extraction RGB model in Stage II can be trained by minimizing the loss as:
\begin{equation} \vspace{-0.05cm}
\mathcal{L}(q, p, n)=\mathcal{L}_{vpr}(\mathbf{g}^{R})+\sum_{\{q, p, n\}}{\mathcal{L}_{kd}(I)}. \vspace{-0.05cm}
\end{equation}






\section{Experiments and Discussions}
In this section, we introduce the experiment details and carry out detailed ablation studies to explore the contribution of each component in StructVPR++. Meanwhile, we compare our method with existing state-of-the-art methods on several standard benchmarks.

\subsection{Experimental Setting}
\label{sec:details}
\begin{table}
\setlength\tabcolsep{1.6pt}
\renewcommand{\arraystretch}{1.2}
\footnotesize
  \caption{\textbf{Summary of Datasets for Evaluation.} $++$, $+$, and $-$ Indicate Degrees from High to Low}
  \label{tbl:dataset}
  \centering
  \scalebox{0.85}{\begin{tabular}{l|c|cc|cccc}
  \toprule
  \multirow{2}{*}{Dataset} &{\textbf{Structural}} & \multicolumn{2}{c|}{Environment}  & \multicolumn{4}{c}{Variation}\\
  \cline{3-8}
   &\textbf{information} & \rotatebox{0}{Urban} & \rotatebox{0}{Suburban} &\rotatebox{0}{View} & \rotatebox{0}{Light} & \rotatebox{0}{Long-term} & \rotatebox{0}{Dynamic} \\
  \midrule
MSLS\cite{msls} & $++$ & \checkmark & \checkmark &  $+$  & $+$  & $+$  & $+$\\
Nordland\cite{nordland} & $+$ & & \checkmark& $-$ & $-$ & $++$  & $-$\\
Pittsburgh\cite{pitts30k}& $-$ &\checkmark && $+$ & $-$ & $-$  & $+$ \\
  \bottomrule                  
  \end{tabular}}
\end{table}

\textbf{Evaluation Datasets.} 
Datasets with still challenging variations are considered in our work, and the summary is shown in \tref{tbl:dataset}, including Mapillary Street Level Sequences (MSLS)\cite{msls}, Pittsburgh\cite{pitts30k}, and Nordland\cite{nordland}.

MSLS~\cite{msls} is introduced to promote lifelong place-recognition research, and contains over 1.6 million images recorded in urban and suburban areas over 7 years.
Compared to other datasets, MSLS covers the most comprehensive variation (dynamic objects, season, light, viewpoint, and weather).
We only evaluate the image-to-image task.
The dataset is divided into a training set, a public validation set and a withheld test set (MSLS challenge)\footnote{\href{https://codalab.lisn.upsaclay.fr/competitions/865}{https://codalab.lisn.upsaclay.fr/competitions/865}}.

Nordland \cite{nordland} contains 4 timestamp-aligned image sequences recorded in 4 seasons.
We use the partitioned dataset \cite{nord1} containing 3450 images per sequence, with summer as reference and winter as query.
Same as \cite{hausler2021patch,nord1}, we remove black tunnels and times when train is stopped.

Pittsburgh \cite{pitts30k} contains 250k images derived from Google Street View panoramas.
The data is generated by 24 perspective images (two pitch and twelve yaw directions) at each place.
In our experiments, we use the subset, Pitts30k, in training, which contains 30k database images and 24k queries, and are divided into train/val./test sets.



\textbf{Evaluation Metrics.}
For all datasets with location labels, {Recall$@N$} is used.
For Recall$@N$, a given query is regarded to be correctly localized if at least one of the top $N$ retrieved database images is within a ground truth tolerance.
Default configurations for these datasets\cite{msls,nord1,pitts30k} are
\begin{itemize}
    \item MSLS: The ground truth corresponding to a query is the reference images located within 25m and 40$^{\circ}$ from the query. In selecting training pairs, we define a distance $d_{qp}$ to represent the FOV overlap between query $q$ and positive $p$:
\begin{equation}
\label{eq:dis}
d_{q p}=\left\|\mathbf{x}_q-\mathbf{x}_p\right\|_2 / 25 + \left(\theta_q-\theta_p\right) / 40 < 1,
\end{equation}
where $\mathbf{x}$ is the GPS coordinate and $\theta$ is the angle. \eref{eq:dis} ensures an overlapping area between a query and a positive.
    \item Nordland: Tolerance is set to 2 frames, that means that one query image corresponds to 5 reference images.
    \item Pittsburgh: Ground truths for evaluation are defined as reference images within 25m from query images.
\end{itemize}

The MSLS Challenge is a hold-out set whose labels are not released, and researchers can submit the predictions on the challenge server to get performance.

\textbf{Training.}
Compared with MSLS, Pittsburgh dataset is smaller and contains many urban buildings shot at close range from non-horizontal perspectives. This results in insufficient categories and instances contained in segmentation images. 
Therefore, in both training stages, we first train models on MSLS, and the performance in urban scenarios (Pitts30k) can be further improved by finetuning on Pitts30k.
For MSLS, images that fit \eref{eq:dis} are selected as positives, and the weakly supervised positive mining strategy proposed in \cite{netvlad} is used on Pitts30k.

In the first learning stage, rgb-branch and seg-branch are trained with the whole backbone with an initial learning rate of 0.001.
The rgb-branch starts with a pre-trained model on ImageNet\cite{krizhevsky2012imagenet}\footnote{\href{https://download.pytorch.org/models/mobilenet_v2-b0353104.pth}{download.pytorch.org/models/mobilenet\_v2-b0353104.pth}} and seg-branch begins with random parameters.
In the second learning stage, backbone also starts with the pre-trained model on ImageNet and is fine-tuned as a whole with an initial learning rate of 0.0001.

Models are all optimized by AdamW optimizer \cite{adamw} with 0.0001 weight decay and cosine learning rate decay schedule, and $m$ in VPR loss is 0.1.
The trained network that yields the best recall@5 on val. set is used for testing.

\textbf{Implementation Details.}
Our method is implemented in PyTorch.
To obtain segmentation images in our proposed framework, we use off-line open-source models (PSPNet\cite{zhao2017pyramid}) and the open-source code-base\footnote{\href{https://github.com/CSAILVision/semantic-segmentation-pytorch}{github.com/CSAILVision/semantic-segmentation-pytorch}} with configuration file of ``ade20k-resnet50dilated-ppm\_deepsup''.


All images are resized to $640 \times 480$.
In the SLME module, StructVPR++ completely abandons the ``dynamic'' category from 6 categories, and $C$ is set as 5, including ``vegetation'', ``sky'', ``ground'', ``buildings'', and ``other objects''.
{For each category, we assign weighted encoding when generating one-hot tensors, with values 0.5, 1, 1, 2, and 2, respectively.}
These settings are derived from human experience.

For {rgb-branch}, we use MobileNetV2 as the backbone, remove the FC layer, and add the MC layer~(\sref{subsec:branch})~(dim=448).
It is also used as the RGB feature extractor in Stage II.
For {seg-branch}, we build a smaller backbone than MobileNetV2 and concatenate the last three levels to conduct global features (dim=480). The backbone refers to the implementation of depth-stream in MobileSal\footnote{\href{https://github.com/yuhuan-wu/MobileSal}{github.com/yuhuan-wu/MobileSal}}.
For group partition strategy, $N_{t}=10$ and $N_{m}=20$.

\subsection{Ablation Studies}
\label{sec:ablation}

We conduct ablative experiments to evaluate the performance of different model variants, illustrating the functionality of each model component, {including the basic components and the newly added components of StructVPR++.}

\subsubsection{Framework}

\textbf{Necessity of Segmentation Images.}
Considering that using segmentation images increases the complexity of training, we first verify the effectiveness of incorporating segmentation images.
Here we compare several other solutions, as shown in \tref{tab:compare_arch}. RGB-i refers to the complex intermediate network structure in \sref{sec:model} combined with only image-level VPR loss.

\begin{table}
  \centering
   \caption{Comparison to Other Fusion Solutions on MSLS. Extraction Latency Includes the Time to Generate Segmentation Images}
  \label{tab:compare_arch}
  \scalebox{0.88}{
  \begin{tabular}{l||ccc||ccc||c}
  \toprule
\multirow{2}{*}{Arch}   & \multicolumn{3}{c||}{MSLS val} & \multicolumn{3}{c||}{MSLS challenge} & Extraction\\
\cline{2-7}
& R@1 & R@5 & R@10 & R@1 & R@5 & R@10 & latency (ms) \\
\midrule
SEG & 72.3 & 82.4 & 85.3 & 48.9 & 64.0 & 70.1 & 8.3$+${385}\\
RGB & 75.8 & 85.3 & 87.3 & 55.1 & 71.9 & 76.4 & 2.25\\
RGB-i & 60.3 & 71.2 & 74.6 & 37.2 & 48.9 & 53.4 & 2.25\\
Multi-task & 75.3 & 86.2 & 88.7 & 56.8 & 72.3& 77.1 & 2.25\\
C-feat & 75.5 & 86.5 & 88.8 & 55.9 & 73.3 & 78.1 & 6.25$+$385\\
C-input & \underline{77.7} & \underline{88.4} & \underline{91.4} & \underline{59.7} & \underline{76.9} & \underline{81.2} & 21.67$+$385\\
\midrule
KD(Ours) & \textbf{84.3} & \textbf{91.5} & \textbf{93.1} & \textbf{65.7} & \textbf{81.4} & \textbf{85.3} & 3.5\\
\bottomrule
\end{tabular}}
\end{table}

\textit{Only using image-level VPR loss:}
On the one hand, it is difficult for the rgb-branch (RGB) in \sref{sec:rgb} to obtain explicit structural information and unified semantic information because it learns more texture-level information under the existing training framework, as shown in \fref{fig:rgb}.
On the other hand, RGB-i performs much worse than RGB because there is no additional supervision to ensure the consistency of feature meanings, causing training to fail to converge or even collapse.

\textit{Multi-task training architecture}: 
Here we use an encoder-decoder structure, where the encoder is shared between two tasks, and the segmentation map is used as the supervision for the decoder.
Note that segmentation is not required during testing. 
We found that the performance has improved under the multi-task framework, which illustrates that segmenting images can bring additional information to VPR.
However, training of Multi-task has high requirements for parameter tuning, and the fusion process is relatively indirect with limited improvement (as shown in \tref{tab:compare_arch}). Therefore, the approach to utilizing segmentation images needs refinement.

\textbf{Comparison with Other Fusion Solutions.}
In the preliminary empirical studies, it has been demonstrated that segmentation images benefit VPR, and many solutions can achieve the fusion of two modalities in global retrieval.
To validate the advantage of knowledge distillation in performance and framework, here we give several feasible solutions and compare them:
\begin{itemize}
    \item Concat-input: Concatenate RGB image and encoded segmentation label map in channel $C$ as input and train the model. Segmentation is required during testing. 
    \item Concat-feat: Two feature vectors from two separate models are concatenated into a global feature. Segmentation is required during testing. 
    \item Knowledge distillation (Ours): 
    Segmentation is not required during testing.
\end{itemize}

{As shown in \tref{tab:compare_arch}}, we train and compare them on MSLS and count the extraction latency during testing. {Extraction latency} is the time to obtain global features from a single RGB image, which includes the generation of segmentation images for SEG, C-feat, and C-input. Compared with feature extraction, the generation of segmentation images takes up more time, so solutions that require online segmentation do not have sufficient real-time performance.

In addition, all fusion algorithms perform better than the two separate branches (RGB, SEG) and Multi-task on MSLS.
But the performance of Concat-feat is limited, which may be due to a lack of coherence between two modalities.
Although both Concat-input and KD perform relatively well, Concat-input requires additional segmentation during testing and has a larger model compared with KD.
Therefore, we use KD in the paper.

\begin{table}[t]
  \centering
\caption{Four Main Components of seg-branch in StructVPR++ Divides for Detailed Ablation Studies}
\label{stab:notation}
  \scalebox{1}{
 \renewcommand{\arraystretch}{1.2}
\begin{tabular}{cc}
\toprule
Notation & Explanation \\
\midrule
$C_1$& Using basic global features\\
$C_2$& Using label features \\
$C_3$& Using weights for label features\\
$C_4$& Using softmax layer for weights\\
\bottomrule
\end{tabular}}
\end{table}

\begin{table}[t]
  \centering
\caption{Comparison of Our Seg-branch and its Variants with Different Combination of Components on MSLS Val. Set}
\label{stab:semanic_abl}
  \scalebox{1}{
\begin{tabular}{c|cccc|ccc}
\toprule
{Method} & $C_1$ & $C_2$ & $C_3$ & $C_4$ & R@1  & R@5 & R@10 \\
\midrule
B    & √ &   &   &   & 69.1 & 80.2& 83.5\\
B-L & √ & √ &   &   & 69.6 & 79.9&83.1\\
B-WP& √ & √ & √ &   & 72.0 &81.6 & 84.1\\
WS  &   & √ & √ & √ & 70.5 &82.3 & 84.7 \\
B-WS(ours) & √ & √ & √ & √ & 72.3 & 82.4 & 85.3 \\
\bottomrule
\end{tabular}}
\end{table}

\subsubsection{Semantic Information Extraction}
\label{subsec:semantic}

\textbf{Design of Label Feature in Seg-branch.}
To explore the performance contribution of each component
in seg-branch, we first divide it into four
components as listed in \tref{stab:notation}.

We study how the proposed label feature generation in seg-branch affects the performance by comparing the standard design (B-WS) with {four degenerate} configurations:
\begin{itemize}
    \item B: Only basic global features $\mathrm{x}^{S}$ are used as final features. 
    \item B-L: Basic global features and unweighted label features are concatenated. In this configuration, we can obtain enhanced global features for seg-branch:
    \begin{equation}
    \label{eq:gsv}
    G^{S} = \mathrm{L2Norm}(\mathrm{Concat}([\mathbf{x}^{S}, \mathbf{l}_{j}^{S}])) \in \mathbb{R}^{(C+1)D_s \times 1}.
    \end{equation}
    \item B-WP: Basic global features and weighed label features are concatenated, but the weights are not normalized by the softmax layer.
    \item WS: Only standard weighted label features are concatenated as the final features.
\end{itemize}

In \tref{stab:semanic_abl}, ablated variants with different combination of these four components are evaluated on MSLS val. set.
There is a performance degradation from the standard seg-branch (B-WS) to B-WP, WS, and then to B-L and B. These results illustrate the importance of using weighted label features for semantic alignment, thereby improving performance.

Firstly, the performance of B-L is similar to B and largely worse than B-WP on the dataset, we believe that's because in the scene where the perspective changes greatly, vanilla label features between the image pair will be misaligned and the way in \eref{eq:gsv} still does not make sense.

\begin{figure}[t]
  \centering
\subfloat[]{
  \label{fig:g}
\includegraphics[width=0.48\columnwidth]{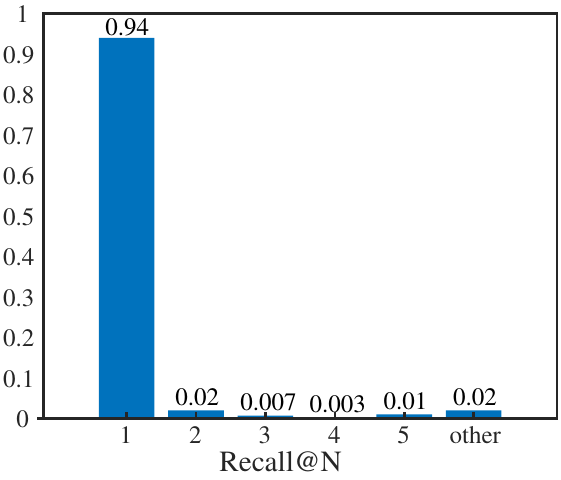}}
\subfloat[]{
  \label{fig:l}
\includegraphics[width=0.48\columnwidth]{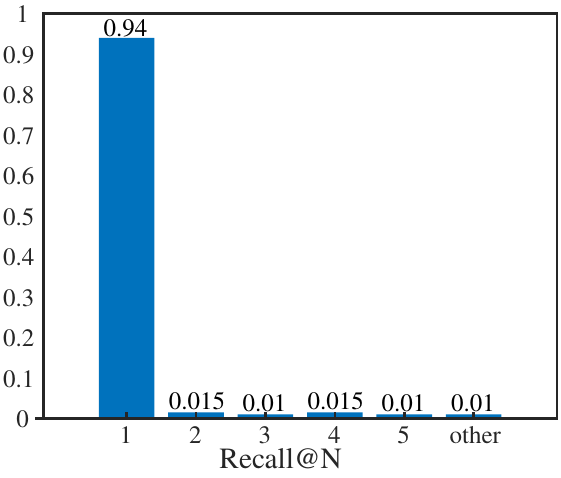}}
   \caption{\textbf{Cross-retrieval ranking of global features and unweighted label features.} After using global features and unweighted category features to rank candidate images respectively, (a) we use unweighted label features to re-rank and count the R@1 candidates for each query image, which are retrieved by global features; (b) we use global features to re-rank and count the R@1 candidates for each query image, which are retrieved by unweighted label features The marked number represents the sample proportion.}
   \label{fig:static}
\end{figure}

To illustrate this argument, we performed additional statistical analysis. 
We use global features and unweighted category features to rank candidate images respectively. For each query image, we then count the ranking of R@1 candidates, which are retrieved by global features, with unweighted label features, and vice versa.
\fref{fig:static} shows that the performance of these two features is very close and the overlap of the retrieved sample rankings is very high.

Secondly, using softmax layer can make the sum of the weights is constrained to be 1, which means attention is allocated to each image, and the relative importance of each label in the image is learned. 

Finally, comparing WS and ours, the structural information contained in basic global features is also important and complementary to semantically aligned information.

\begin{table}[t]
  \centering
\caption{Comparison of Our Label-aware RGB Model in Stage II and its Variants on Datasets}
\label{stab:rgb_abl}
  \scalebox{0.82}{
 \renewcommand{\arraystretch}{1.2}
\begin{tabular}{c|ccc|ccc|ccc}
\toprule
\multirow{2}{*}{Method} & \multicolumn{3}{c|}{MSLS val}& \multicolumn{3}{c|}{MSLS test} & \multicolumn{3}{c}{Nordland} \\
\cmidrule(lr){2-10}
&R@1  & R@5  & R@10 &R@1  & R@5  & R@10  &R@1  & R@5  & R@10\\
\midrule
ours-G & 83.2&91.2&92.8&64.8&80.9&84.4&56.7&76.1&83.2\\
\textbf{ours} &{84.3}&91.5&93.1&65.7&81.4&{85.3} & 58.4&77.3&84.1\\
\bottomrule
\end{tabular}}
\end{table}

\textbf{Usage of Label Features in Label-aware RGB Model.}
In addition to seg-branch, label features are used throughout the whole framework. To verify the effectiveness of label features during testing, we trained and evaluated StructVPR++ with only basic global RGB features in Stage II (Stage I is same to our proposed framework), labeled as \textit{ours-G}. 
While using only basic global features, the architecture of final model is same as rgb-branch, and only a MLP layer is used to align enhanced global SEG features and RGB features into a same domain.
As shown in {\tref{stab:rgb_abl}}, the standard StructVPR++ achieves better results than the degenerate version.
On the one hand, the results show that as long as the training strategy is appropriate, label-aware RGB model can also be improved with the help of weighted label features.
On the other hand, the results illustrate that knowledge for semantic alignment can be learned in label-aware RGB model through the proposed KD-based framework.






\subsubsection{Group Partition and Weighting Distillation}

\begin{figure}[t]
  \centering
   \includegraphics[width=0.98\linewidth]{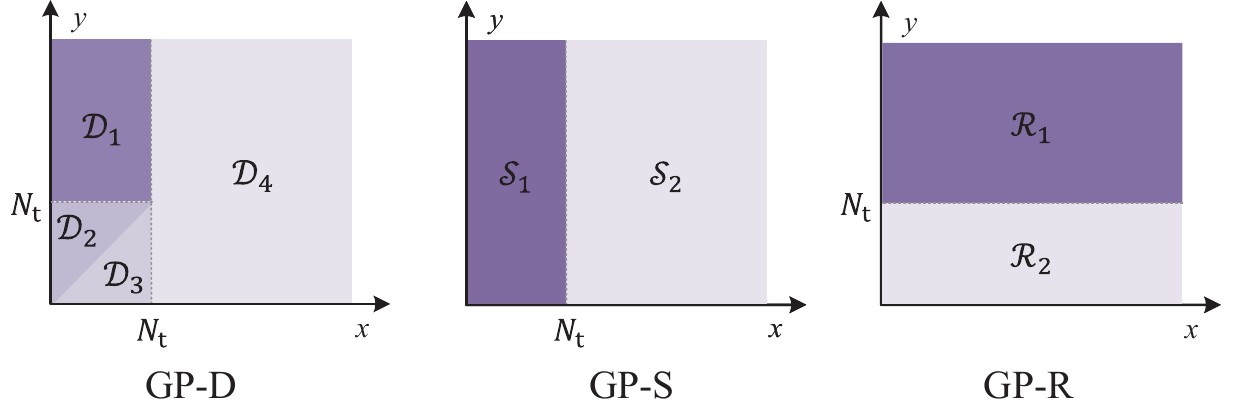}
   \caption{\textbf{Visualization of different group partition strategies.} Compared with our strategy (GP-D), which uses two models, both GP-S and GP-R have only one model involved in the partition.}
   \label{fig:group}
\end{figure}

As shown in \fref{fig:group}, GP-D refers to our group partition strategy in \sref{subsec:group}.
This paper proposes that not all samples are helpful for distillation and uses group partition to select samples.
{To prove this insight and motivation}, we first present the experiments on impacts of different groups.
For convenience, we then present the ablation studies on group partition strategy with the setting of selective knowledge distillation.
Finally, we carry out ablation experiments on weighting distillation loss with two partition strategies.

\textbf{Impact of Groups.}
In \tref{tab:group_abl}, we sequentially select a group from $\{ \mathcal{D}_1, \mathcal{D}_2, \mathcal{D}_3, \mathcal{D}_4 \}$ to participate in distillation and evaluate the effectiveness. 
\textbf{None} refers to the {rgb-branch} without distillation and \textbf{All} stands for non-selective KD using all samples for KD.
Compared with \textbf{None}, it can be concluded that ${\mathcal{D}_{4}}$ will harm distillation and other groups have different degrees of positive effects.
Moreover, the results of \textbf{All} are not as good as ${\mathcal{D}_{1}}$, indicating that the quality of samples is as important as the quantity for KD.
Therefore, it is necessary to select appropriate samples.

\begin{table}[t]
  \centering
\caption{Performance of Selective Distillation using {Basic Global Features} with Different Samples. \textit{None} Refers to the {Rgb-branch} without Distillation and \textit{All} Refers to Non-selective Distillation}
\label{tab:group_abl}
  \scalebox{0.93}{
\begin{tabular}{c|c|ccc|ccc}
\toprule
\multirow{2}{*}{\begin{tabular}[c]{@{}c@{}}Group for \\ distillation\end{tabular}} & \multirow{2}{*}{\begin{tabular}[c]{@{}c@{}} Sample\\ratio \end{tabular}}&\multicolumn{3}{c|}{MSLS val} & \multicolumn{3}{c}{MSLS challenge} \\
\cmidrule(lr){3-8}
& & R@1      & R@5     & R@10    & R@1       & R@5       & R@10      \\
\midrule
None & 0\% & 75.8 & 85.3& 87.3 & 55.1 & 71.9 & 76.4\\
\textbf{All} & 100\% &78.4 & 87.4& 90.1 & 59.1 & 73.5& 79.3\\ 
\midrule
${\mathcal{D}_{1}}$ & 4.36\% & 78.8 & \underline{89.2}& 91.1 & \underline{62.3} & 76.8 & 80.9\\ 
${\mathcal{D}_{2}}$ & 50.62\%  & 80.0 & 88.8& 90.6 & 59.1&75.5&79.3\\ 
${\mathcal{D}_{3}}$ & 14.87\%  &77.7 & 87.4 & 89.2 & 57.8&74.4&79.1 \\ 
${\mathcal{S}_{1}}$& 67.75\% & \underline{81.6} & 88.9 & \underline{91.2} & {62.2} & \underline{77.2} & \underline{82.0}\\ 
${\mathcal{S}_{2}}({\mathcal{D}_{4}})$ & 32.25\% & 71.6&83.7&85.1 & 48.2 & 63.9 & 68.7 \\ 
${\mathcal{R}_{1}}$ & 73.95\% &79.3&87.6&88.9 & 57.0&74.1&78.2\\ 
${\mathcal{R}_{2}}$ & 26.05\% &73.8&82.7&85.6 & 51.0&67.0&72.3\\ 
\midrule
\textbf{Ours-global} & 69.55\% & \textbf{83.0} & \textbf{91.0} & \textbf{92.6} & \textbf{64.5} & \textbf{80.4}& \textbf{83.9}\\
\bottomrule
\end{tabular}}
\end{table}

\textbf{Group Partition Strategy.}
In this experiment, for convenience, we only perform \textit{selective knowledge distillation}, which is a degenerate version of weighted KD.
To validate the advantage of our group partition strategy, we compare our standard partition strategy (GP-D) with two degenerate configurations that use a single branch, as shown in \fref{fig:group}:
\begin{itemize}
    \item \textit{Group partition with seg-branch}  (GP-S). Samples in $\mathcal{S}_1$ are those seg-branch (teacher) is good at; samples in $\mathcal{S}_2$ are those seg-branch is not good at and should be ignored in distillation.
    \item \textit{Group partition with rgb-branch} (GP-R). Samples in $\mathcal{R}_1$ are those rgb-branch (student) is good at; samples in $\mathcal{R}_2$ are those rgb-branch is good at and should be excluded from distillation.
\end{itemize}

{For selective KD, we exclude the samples with zero weight.
In other words, GP-D uses samples belonging to $\{{\mathcal{D}_{1}}, {\mathcal{D}_{2}}, {\mathcal{D}_{3}}\}$, GP-S uses $\{{\mathcal{S}_{1}}\}$, and GP-R uses $\{{\mathcal{R}_{1}}\}$.}
In this experiment, GP-D degenerates to GP-S, and GP-S performs better than GP-R, as shown in \tref{tab:group_abl}.

\begin{table}[t]
  \centering
\caption{Performance of Weighted Distillation with Different Weights. The Weights Correspond to $\mathcal{S}_{1}$-$\mathcal{S}_{2}$ and $\mathcal{D}_{1}$-$\mathcal{D}_{2}$-$\mathcal{D}_{3}$-$\mathcal{D}_{4}$, and GP-S(1-0) is Equal to GP-D(1-1-1-0)}
\label{tab:weight_abl}
  \scalebox{1}{
\begin{tabular}{c|ccc|ccc}
\toprule
\multirow{2}{*}{\begin{tabular}[c]{@{}c@{}}Weight\end{tabular}} & \multicolumn{3}{c|}{MSLS val} & \multicolumn{3}{c}{MSLS challenge} \\
\cline{2-7}
& R@1      & R@5     & R@10    & R@1       & R@5       & R@10      \\
\midrule
GP-S(1-0) &  {82.0} & 89.1 & 91.5 & {62.6} & 77.7 & 82.2\\
GP-D(8-4-1-0) & \underline{82.3} & 89.1 & 91.7 & {62.5} &\underline{79.3} & 82.5\\ 
\midrule
GP-S($\varphi'(q,p)$)& 81.5 & \underline{89.8} & \underline{92.4} & \underline{63.1} & {78.8} & \underline{83.9}\\ 
GP-D($\varphi(q,p)$) & \textbf{84.3} & \textbf{91.5} & \textbf{93.1} & \textbf{65.7} & \textbf{81.4} & \textbf{85.3} \\
\bottomrule
\end{tabular}}
\end{table}

\begin{table*}[t]
  \centering
   \caption{Comparison to SOTA Methods on Several Standard Datasets. The Results of Global Retrieval and Re-ranking with Local Features are Shown}
  \label{tab:compare_SOTA}
  \scalebox{0.95}{
  \renewcommand{\arraystretch}{1.2}
  \begin{tabular}{ll||ccc||ccc||ccc||ccc@{}}
  \toprule
 &\multirow{2}{*}{Method} & \multicolumn{3}{c||}{MSLS val} & \multicolumn{3}{c||}{MSLS challenge} & \multicolumn{3}{c||}{Nordland test} & \multicolumn{3}{c}{Pittsburgh30k test}\\
\cline{3-14}
& & R@1 & R@5 & R@10 & R@1 & R@5 & R@10 & R@1 & R@5 & R@10  & R@1 & R@5 & R@10 \\
\hline
\hline
\multirow{9}{*}{\begin{tabular}[c]{@{}c@{}}Global\\ retrieval\end{tabular}} &
NetVLAD\cite{netvlad}&53.1 & 66.5 & 71.1 &28.6 & 38.3 & 42.9 & 11.5 & 17.6&  21.9 & 81.9 & 91.2 & 93.7\\
& SFRS\cite{ge2020self}&58.8 & 68.2 & 71.8 & 30.7 & 39.3 & 42.9 & 14.3 & 23.1 & 27.3 & {71.1} & {81.0} & {84.9}\\
& DELG\cite{delg}&68.4&78.9 &83.1 & 37.6& 50.5& 54.6 & 27.0&43.3&50.0 & 79.0 & 89.0 & 92.7 \\
& Patch-NetVLAD-s\cite{hausler2021patch}&63.5 & 76.5 & 80.1 & 36.1 & 49.9 & 55.0 &  18.1 & 33.2  & 41.1 & 81.3 & 91.1& 93.4 \\
& Patch-NetVLAD-p\cite{hausler2021patch}&70.0 & 80.4 & 83.8 & 38.1 & 51.2 & 55.3 & 24.8 & 39.4 & 48.0 & \underline{83.7} & \underline{91.8} & \underline{94.0}\\
& TransVPR \cite{transvpr}&{70.8}&{85.1} & {89.6}  & {48.0} & {67.1} & {73.6} & {31.3} & {53.6} & {64.8} & 73.8 & 88.1 & 91.9\\
& R2Former\cite{r2former} &\underline{79.3}&\underline{90.8}&\underline{92.6}&\underline{57.2}&\underline{76.0}&\underline{81.7}&\underline{35.3}&\underline{57.6}&\underline{67.6}&77.7&90.5&93.5\\
& StructVPR\cite{structvpr} &{83.0}&{91.0}&{92.6}&{64.5}&{80.4}&{83.9}&{56.1}&{75.5}&{82.9}&{85.1}&{92.3}&{94.3}\\
\cline{3-14}
& \textbf{(i) Ours} &\textbf{84.3} & \textbf{91.5} & \textbf{93.1} & \textbf{65.7} & \textbf{81.4}& \textbf{85.3} & \textbf{58.4} & \textbf{77.3} & \textbf{84.1} & \textbf{85.4} & \textbf{92.5} & \textbf{94.5}\\
\hline
\hline
\multirow{9}{*}{Re-ranking} &
SP-SuperGlue\cite{detone2018superpoint,sarlin2020superglue}&78.1 & 81.9 & 84.3 & 50.6 & 56.9 & 58.3 & 37.9 & 41.2 & 42.6 & 87.2 & 94.8 & 96.4\\
& DELG\cite{delg}& 83.9 & 89.2 & 90.1 & 56.5 & 65.7 & 68.3 &64.4&70.8&72.7& {89.9} & \underline{95.4} & \underline{96.7} \\
& Patch-NetVLAD-s\cite{hausler2021patch}&77.2 & 85.4 & 87.3 & 48.1 & 59.4 & 62.3 & 50.9 & 62.7 & 66.5 & 88.0 & 94.5 & 95.6 \\
& Patch-NetVLAD-p\cite{hausler2021patch}&79.5 & 86.2 & 87.7 & 51.2 & 60.3 & 63.9 & 62.7 & 71.0 & 73.5 & 88.7 & 94.5 & 95.9 \\
& TransVPR \cite{transvpr}& {86.8} & 91.2 & 92.4 & 63.9 & 74.0 & 77.5 & \underline{77.8} & {86.8} & 89.3 & 89.0 & 94.9 & 96.2\\
& R2Former\cite{r2former} &\textbf{89.7}&\textbf{95.0}&\underline{96.2}&\textbf{73.0}&\textbf{85.9}&\textbf{88.8}&77.1&\underline{89.0}&\underline{91.6}&\textbf{91.1}&95.2&96.3\\
& StructVPR-SuperGlue\cite{structvpr,sarlin2020superglue}&88.4&94.3&95.0&69.4&81.5&85.6&{83.5}&{93.0}&{95.0}&90.3&{96.0}&{97.3}\\
\cline{3-14}
& \textbf{(ii) Ours-SP-SuperGlue}&\underline{89.5}&\underline{94.8}&\textbf{96.3}&\underline{72.5}&\underline{85.1}&\underline{87.9}&\textbf{84.6}&\textbf{93.8}&\textbf{95.6}&\underline{90.7}&\textbf{96.8}&\textbf{97.8}\\
\bottomrule
\end{tabular}}
\end{table*}

\textbf{Weighting distillation loss.}
To validate the effect of weighting distillation loss, we compare our weighting function (\eref{eq:rule}) with constant weights to illustrate the importance of fine-grained weight assignment for samples.
We have verified the importance of the groups above, which guides the design \eref{eq:rule}: the more essential samples for distillation, the higher the weight.

To further illustrate the advantages of group partition and GP-D strategy, we also compare GP-D and GP-S with weighting functions.
For GP-S, \eref{eq:rule} degenerates as follows:
\begin{equation}
\label{eq:rule1}
\varphi'(q,p)=\\
\begin{cases}
1+\frac{1}{4 \cdot \ln \left(1+RN_{s}\right)}, &(q,p) \in {\mathcal{S}_{1}}\\
0, &(q,p) \in {\mathcal{S}_{2}}
\end{cases},
\end{equation}
where the performance gap between the student network and the teacher network is not captured in the sample-specific weighting.

In \tref{tab:weight_abl}, GP-D(8-4-1-0) performs better than GP-S(1-0), and GP-D($\varphi(q,p)$) is better than GP-S($\varphi'(q,p)$), which illustrate the importance of precise partition.
Moreover, for both GP-D and -S, the weight function performs better than the discrete constant weights, showing the advantages of setting specific weights for each sample.

\subsection{Performance Comparison With State-of-the-Arts}

\textbf{Baselines.} We compared our method against several state-of-the-art image retrieval-based localization solutions, 
including methods using global descriptors only:
\textbf{NetVLAD}\cite{netvlad}and \textbf{SFRS}\cite{ge2020self}, and methods which additionally perform re-ranking using spatial verification of local features: \textbf{DELG}\cite{delg},
\textbf{Patch-NetVLAD}\cite{hausler2021patch}, \textbf{SP-SuperGlue}, \textbf{TransVPR}\cite{transvpr}, and \textbf{R2Former}\cite{r2former}.
DELG, Patch-NetVLAD, TransVPR, and R2Former jointly extract global and local features for image retrieval, while SP-SuperGlue re-ranks NetVLAD retrieved candidates by using SuperGlue\cite{sarlin2020superglue} matcher to match SuperPoint\cite{detone2018superpoint} local features. 

Here we give the key details of the reproduction of some SOTA algorithms.

For \textbf{NetVLAD} \footnote{\href{https://github.com/Nanne/pytorch-NetVlad}{github.com/Nanne/pytorch-NetVlad}} , we select the released model trained on Pitts30k training set with VGG-16 backbone.

For \textbf{SFRS} \footnote{\href{https://github.com/yxgeee/OpenIBL}{github.com/yxgeee/OpenIBL}}, we use the released model trained on Pitts30k training set.

For \textbf{Patch-NetVLAD}\footnote{\href{https://github.com/QVPR/Patch-NetVLAD}{github.com/QVPR/Patch-NetVLAD}}, we tested its speed-focused and performance-focused configurations, denoted as Patch-NetVLAD-s and Patch-NetVLAD-p, respectively.

For \textbf{DELG}\footnote{\href{https://github.com/feymanpriv/DELG}{https://github.com/feymanpriv/DELG}}, we change the extraction of global features as the original paper. For local features, all the reproduced results are from the provided model (dim=512), which is different from the original paper (dim=128).

For \textbf{SP-SuperGlue}\footnote{\href{https://github.com/magicleap/SuperGluepre-trainedNetwork}{github.com/magicleap/SuperGluepre-trainedNetwork}}, our reproduction includes: using NetVLAD for global retrieval, then extracting SuperPoint local features, and applying SuperGlue to identify matches and to re-rank candidates. We choose the pre-trained outdoor weights on MegaDepth dataset\cite{li2018megadepth}. For \textbf{TransVPR} \footnote{\href{https://github.com/RuotongWANG/TransVPR-model-implementation}{github.com/RuotongWANG/TransVPR-model-implementation}} and \textbf{R2Former}\footnote{\href{https://github.com/bytedance/R2Former.git}{github.com/bytedance/R2Former.git}}, we use the pretrained models.

{\textbf{Re-ranking Backend.}}
Candidates obtained through global retrieval are often re-ranked using geometric verification or feature fusion.
In this paper, our main contribution is based on extracting global features, so we use the classic re-ranking methods (i.e., SuperGlue) as the backend to compare with the other two-stage algorithms.
SuperPoint descriptors are used as our local features.
For SuperGlue, we use the official implementation and configuration.

\textbf{Quantitative comparison.}
\tref{tab:compare_SOTA} compares ours against the retrieval approaches.
There are two settings for StructVPR++: (i) global feature similarity search and (ii) global retrieval followed by re-ranking with local feature matching (SP-SuperGlue).
In the experiments, we first compare our global model with all methods. Then re-ranking algorithms are used as our backend to compare with those two-stage algorithms.
The top 100 images are re-ranked. 

In setting (i), ours-global convincingly outperforms all compared global methods on MSLS validation, MSLS challenge, Nordland, and Pittsburgh30k datasets. Compared with the best global baseline, {the absolute increments on Recall$@1$ are 5$\%$, 8.5$\%$, 23.1$\%$, and 7.7$\%$, and the ones on Recall$@5$ are 0.7$\%$, 5.4$\%$, 19.7$\%$, and 2$\%$ respectively.}
More importantly, ours-global is better than almost all two-stage ones on Recall$@5$ and Recall$@10$, which shows the potential of our method for improvement on Recall$@1$.
Ours-global also achieves SOTA results on Pitts30k dataset, with a 1.7$\%$ increment on Recall$@1$ over Patch-NetVLAD.
{We observe that the global model trained on MSLS shows a significant improvement compared with previous SOTA methods and our conference version, while the one by Pitts is relatively mediocre. Such results are due to differences in datasets, that is, whether the scene contains enough structural information and semantic alignment knowledge.
This experimental phenomenon also proves that datasets do not limit StructVPR++: on datasets with obvious structural information, it can effectively improve performance; on datasets with few structural information, it can achieve effective distillation without hurting the RGB performance.}

In setting (ii), compared with SP-SuperGlue, the results of ours-SP-SuperGlue illustrate the importance of global retrieval.
In addition, ours-SP-SuperGlue achieves SOTA performance {on Nordland and Pitts30k datasets} by an average of 3.2\% (absolute R@$5$ increase) and by 2.75\% (absolute R$@5$ increase), while a little lower than R2Former on MSLS datasets.
We think this is because global features and local features are not trained together.

\begin{table}[b]
  \centering
\caption{\textbf{Latency and Memory Footprint.} The Following Data is Measured on NVIDIA GeForce RTX 3090 GPU and Intel Xeon Gold 6226R CPU. For Global Retrieval Methods, Matching Time and Memory are Negligible}
\scalebox{0.85}{
\begin{tabular}{cl|c|c|c}
\toprule
& Method & \begin{tabular}[c]{@{}c@{}}Extraction\\ latency (ms)\end{tabular} &         \begin{tabular}[c]{@{}c@{}}Matching\\ time (s)\end{tabular} & \begin{tabular}[c]{@{}c@{}}Memory \\ (MB)\end{tabular} \\
\midrule
\multirow{3}{*}{\begin{tabular}[c]{@{}c@{}}Global\\ retrieval\end{tabular}}
& NetVLAD\cite{netvlad} & 40 & $-$ & $-$ \\
& SFRS\cite{ge2020self} & 207 & $-$ & $-$ \\
& Ours & 3.5 & $-$ & $-$\\
\midrule
\multirow{7}{*}{Re-ranking}
& SP-SuperGlue\cite{detone2018superpoint,sarlin2020superglue} & 35 & 6.4 & 1.93 \\
& DELG\cite{delg} & 199 & 45.7 & 0.37 \\
& Patch-NetVLAD-s\cite{hausler2021patch} & 42.5 & 3.29 & 1.82 \\
& Patch-NetVLAD-p\cite{hausler2021patch} & 625 &  25.6 & 44.14 \\
& TransVPR\cite{transvpr} & 8 & 3.5 & 1.17 \\
& R2Former\cite{r2former} & 11.4 & 0.5 & 0.24 \\
& Ours-SP-SuperGlue & 3.5 & 3.9 &1.47\\
\bottomrule
\end{tabular}
}
\label{tab:latency_memory}
\end{table}

\textbf{Latency and memory footprint.}
In real-world VPR systems, latency and resource consumption are important factors.
As shown in \tref{tab:latency_memory}, latency and memory requirements refer to processing a single query image. All the methods are measured on MSLS Val (18, 871 database images).

Ours has a great advantage in extraction latency over all other algorithms.
In addition, Ours-SP-SuperGlue achieves SOTA recall performance on average, and it is 11.7 times and 6.6 times faster than DELG and Patch-NetVLAD-p in spatial matching.

In summary, previous high-accuracy two-stage methods mainly rely on re-ranking, which comes at a high computational cost. On the contrary, our global retrieval results are good with low computational cost, achieving a better balance of accuracy and computation.

\subsection{Qualitative Results}

\begin{figure}[t]
    \centering
    \includegraphics[width=0.93\linewidth]{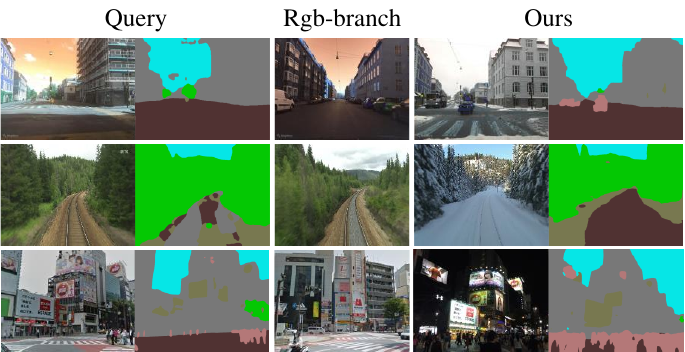}
    \caption{In these examples, the texture information among images at the same location is dissimilar, and StructVPR++ successfully retrieves matching images, while rgb-branch produces incorrect place matches.}
    \label{fig:qplot}
\end{figure}

\textbf{Effect of Using Segmentation.}
Here, we present qualitative examples on MSLS in \fref{fig:qplot} to demonstrate our framework can improve the robustness of the algorithm to scenarios where texture information is dissimilar and affected by light and seasonal changes.
For rgb-branch, we can find that the model pays more attention to texture information, such as the color of the image. In contrast, StructVPR++ will combine texture information and structural information.

\begin{figure}[t]
    \centering
    \includegraphics[width=0.87\linewidth]{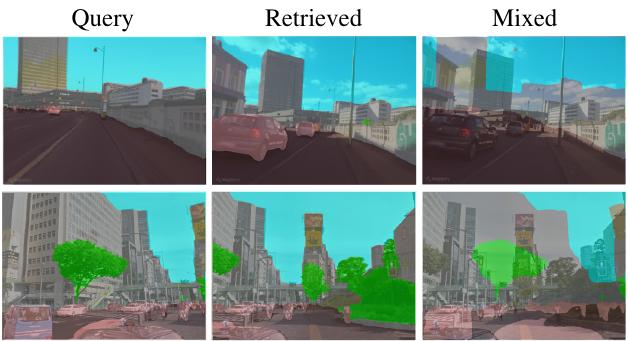}
    \caption{In these examples, the co-viewing angle between the query image and the GT candidate image is small, and the right column shows the superposition of the query segmentation and the retrieved images. Although the structural information similarity between the two is very low, our final seg-branch (B-WS) successfully retrieves GT images.}
    \label{fig:splot}
\end{figure}

\textbf{Effect of Weighted Label Features.}
As shown in \fref{fig:splot}, the co-viewing angle between the query and the GT candidate is small.
Same as to \sref{subsec:semantic}, we test our seg-branch (B-WS) and its variant model (B) with $C=5$ on these examples.
Before adding weighted label features, the model has limited ability to handle scenes with large viewing angle changes.
In contrast, StructVPR++ can independently select the main label of the scene for matching.

More qualitative results are shown in the Supplementary Material.

\subsection{Discussions}
For strengths, our method is capable of achieving the combination of SEG and RGB while avoiding segmentation in testing through the distillation framework. In addition, the proposed feature extraction module can explicitly decouple label features to solve semantic misalignment. At the same time, the different effects of samples are effectively utilized to distill high-quality knowledge. More importantly, our method can achieve SOTA performance among all global methods and even achieve comparable performances to most two-stage methods.

{For limitations and future work, some challenges remain. The weighting function is defined through a heuristic way, and it is worth exploring whether learning methods would be better to obtain the weights through joint optimization.
Moreover, better backbones with stronger representation capabilities can be used for RGB feature extraction to improve final absolute performance.
Furthermore, we believe that StructVPR++ is applicable to other tasks or modalities as long as multi-modal information is complementary at the sample level for tasks.
}


\section{Conclusion}
\label{sec:conclusion}
In this paper, we propose StructVPR++ for visual place recognition, a distillation-driven framework, {which transfers structural and semantic cues from segmentation images to RGB global representations.}
Firstly, we incorporate domain knowledge to encode segmentation images into weighted one-hot format with coarse-grained labels.
{Secondly, we introduce a label-aware feature decoupling mechanism that isolates semantic-specific patterns from global descriptors.}
Thirdly, we propose the group partition strategy and a sample-based weighting function for distillation, thereby enhancing beneficial knowledge in RGB features.
Experiments fairly demonstrate the advancement of the proposed framework, and our global retrieval achieves a better trade-off between accuracy and efficiency.

\section*{Acknowledgments}
This work was supported by the National Natural Science Foundation of China under Grants 62088102.

\ifCLASSOPTIONcaptionsoff
  \newpage
\fi

\bibliographystyle{IEEEtran}
\bibliography{ref}

\begin{IEEEbiography}[{\includegraphics[width=1in,height=1.25in,clip,keepaspectratio]{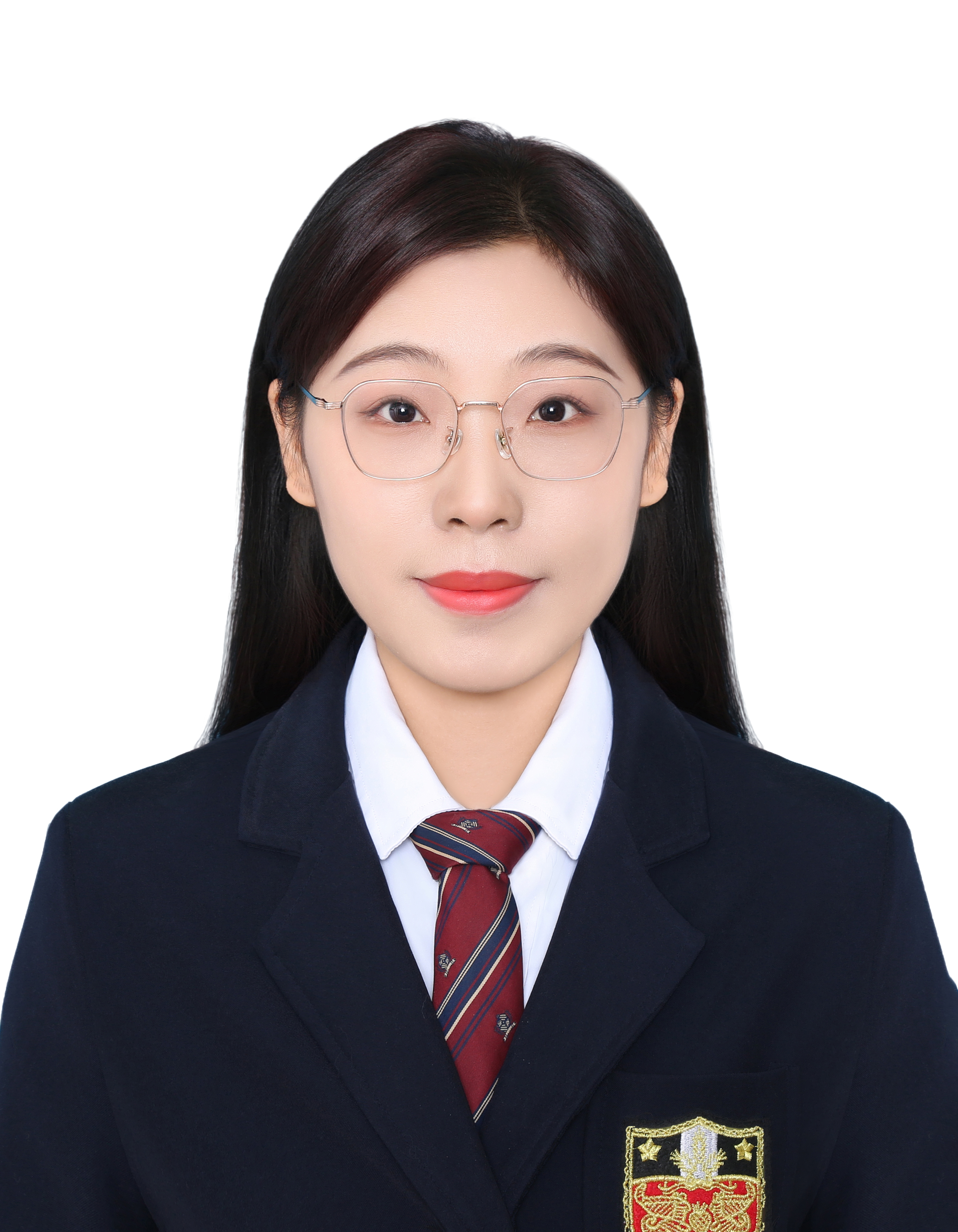}}]{Yanqing Shen}
received the B.E. degree in electronic and information engineering from Xi’an Jiaotong University, Xi'an, China, in 2019, where she is currently pursuing the Ph.D. degree with the Institute of Artificial Intelligence and Robotics.
Her research interests mainly include visual place recognition, autonomous vehicle localization, and embodied AI.
She has published several papers in CVPR, RA-L, ICRA, IROS, and so on.
\end{IEEEbiography}

\vspace{1pt}

\begin{IEEEbiography}[{\includegraphics[width=1in,height=1.25in,clip,keepaspectratio]{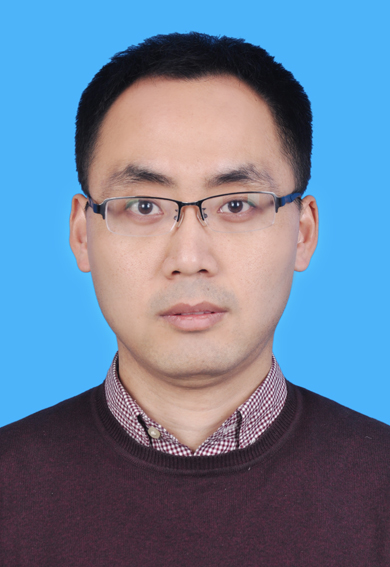}}]{Sanping Zhou}
(Member, IEEE) received the Ph.D. degree in control science and engineering from Xian Jiaotong University, Xi’an, China, in 2020. From 2018 to 2019, he was a Visiting Ph.D. Student with the Robotics Institute, Carnegie Mellon University. He is currently an Associate Professor with the Institute of Artificial Intelligence and Robotics, Xian Jiaotong University. His research interests include machine learning, deep learning, and computer vision, with a focus on person re-identification, medical image segmentation, image classification, and visual tracking.
\end{IEEEbiography}

\vspace{1pt}

\begin{IEEEbiography}[{\includegraphics[width=1in,height=1.25in,clip,keepaspectratio]{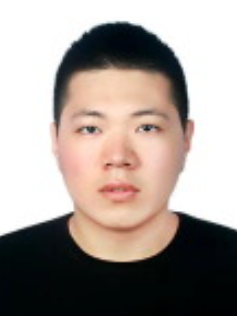}}]{Jingwen Fu}
received the B.S. degree from the School of Electronic and Information Engineering, Xi’an Jiaotong University, Xi’an, China, in 2021. He is presently a doctoral candidate at the same institution. His research is centered on deciphering the learning processes of deep learning models and leveraging these insights to develop more efficient artificial intelligence systems.
\end{IEEEbiography}

\vspace{1pt}

\begin{IEEEbiography}[{\includegraphics[width=1in,height=1.25in,clip,keepaspectratio]{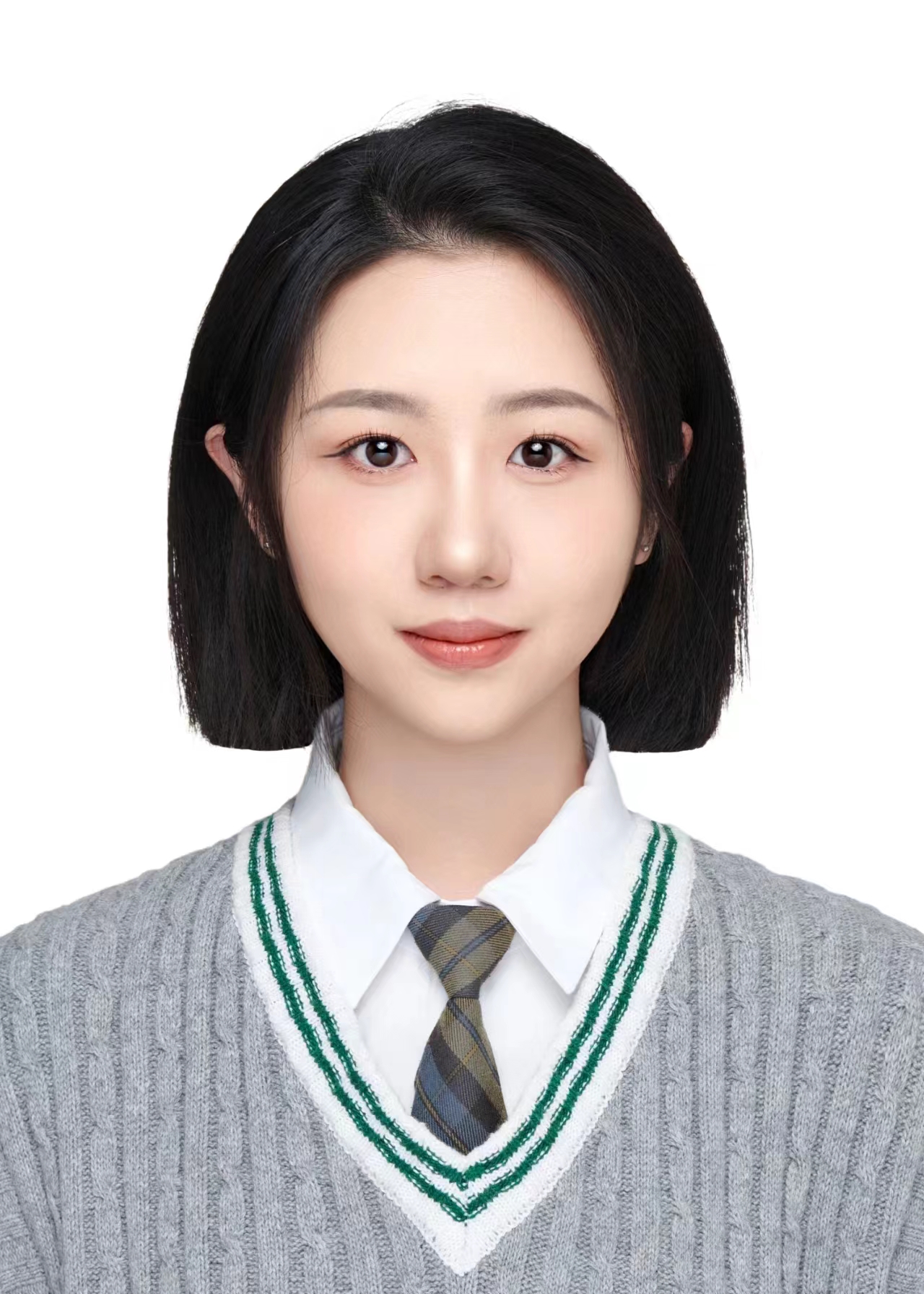}}]{Ruotong Wang}
received the B.E. degree in electronic and information engineering and the M.S. degree in Control Science and Engineering from Xi’an Jiaotong University, Xi'an, China, in 2020 and 2023, respectively.
Her research interests mainly include visual place recognition and object detection.
\end{IEEEbiography}

\vspace{1pt}

\begin{IEEEbiography}[{\includegraphics[width=1in,height=1.25in,clip,keepaspectratio]{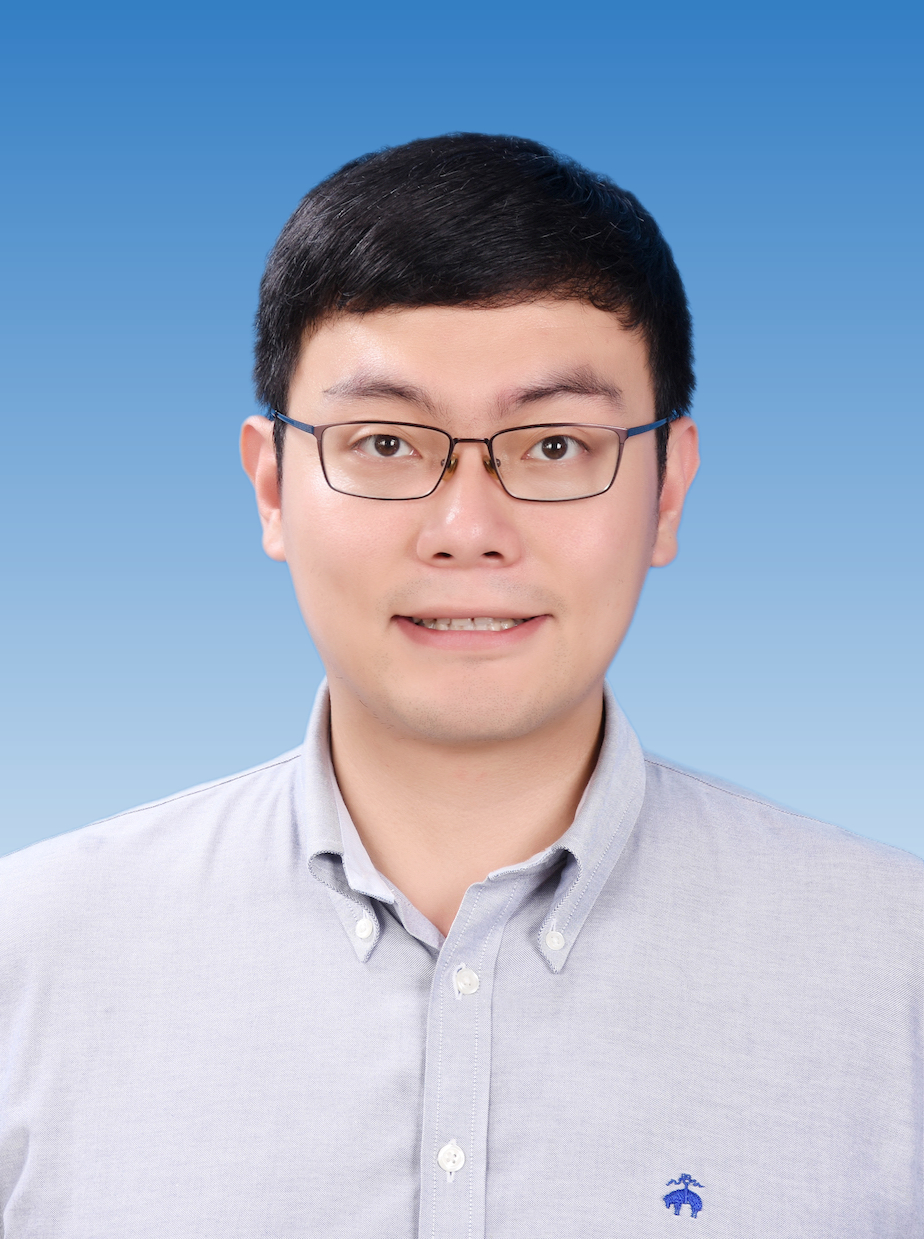}}]{Shitao Chen} (Member, IEEE) received his Ph.D. degree in Electronic and Information Engineering from Xi'an Jiaotong University in 2022. He is currently serving as an assistant professor at the Institute of Artificial Intelligence and Robotics at Xi'an Jiaotong University. His research interests include computer vision, pattern recognition, deep learning, image processing, and hardware implementation of intelligent systems.
\end{IEEEbiography}

\vspace{1pt}

\begin{IEEEbiography}[{\includegraphics[width=1in,height=1.25in,clip,keepaspectratio]{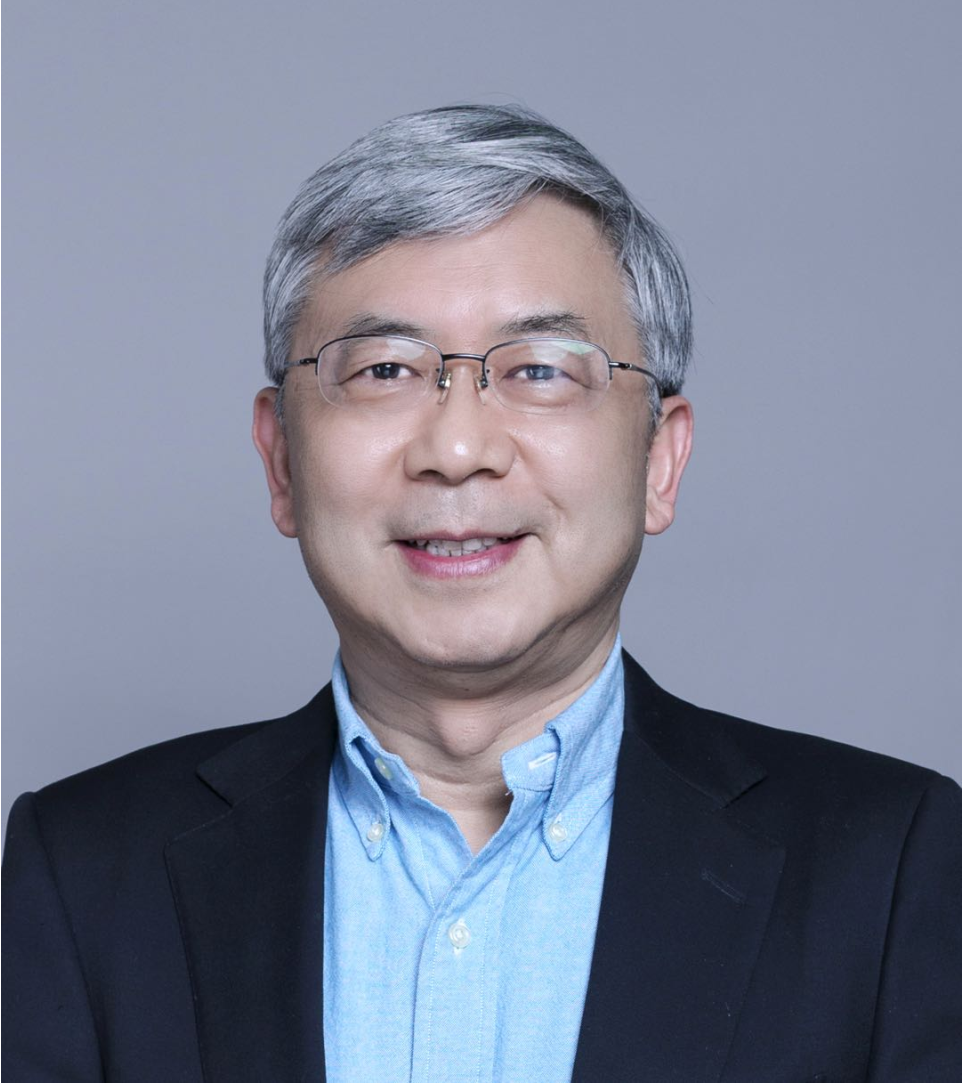}}]{Nanning Zheng}
(Fellow, IEEE) graduated in
1975 from the Department of Electrical Engineering, Xi’an Jiaotong University (XJTU), received the M.E. degree in Information and Control Engineering from Xi’an Jiaotong University in 1981, and a Ph.D. degree in Electrical Engineering from Keio University in 1985. He is currently a Professor and the Director with the Institute of Artificial Intelligence and Robotics of Xi’an Jiaotong University. His research interests include computer vision, pattern recognition, and hardware implementation of intelligent systems. Since 2000, he has been the Chinese representative on the Governing Board of the International Association for Pattern Recognition. He became a member of the Chinese Academy Engineering in 1999. He is a fellow of the IEEE.
\end{IEEEbiography}

\end{CJK*}
\end{document}


\begin{CJK*}{UTF8}{gbsn}
\CJKindent

\title{StructVPR++: Distill Structural and Semantic Knowledge with Weighting Samples for Visual Place Recognition\\Supplementary Material}

\maketitle



\begin{figure}[t]
    \centering
    \subfloat[]{
      \label{fig:seg_val}
    \includegraphics[width=0.48\linewidth]{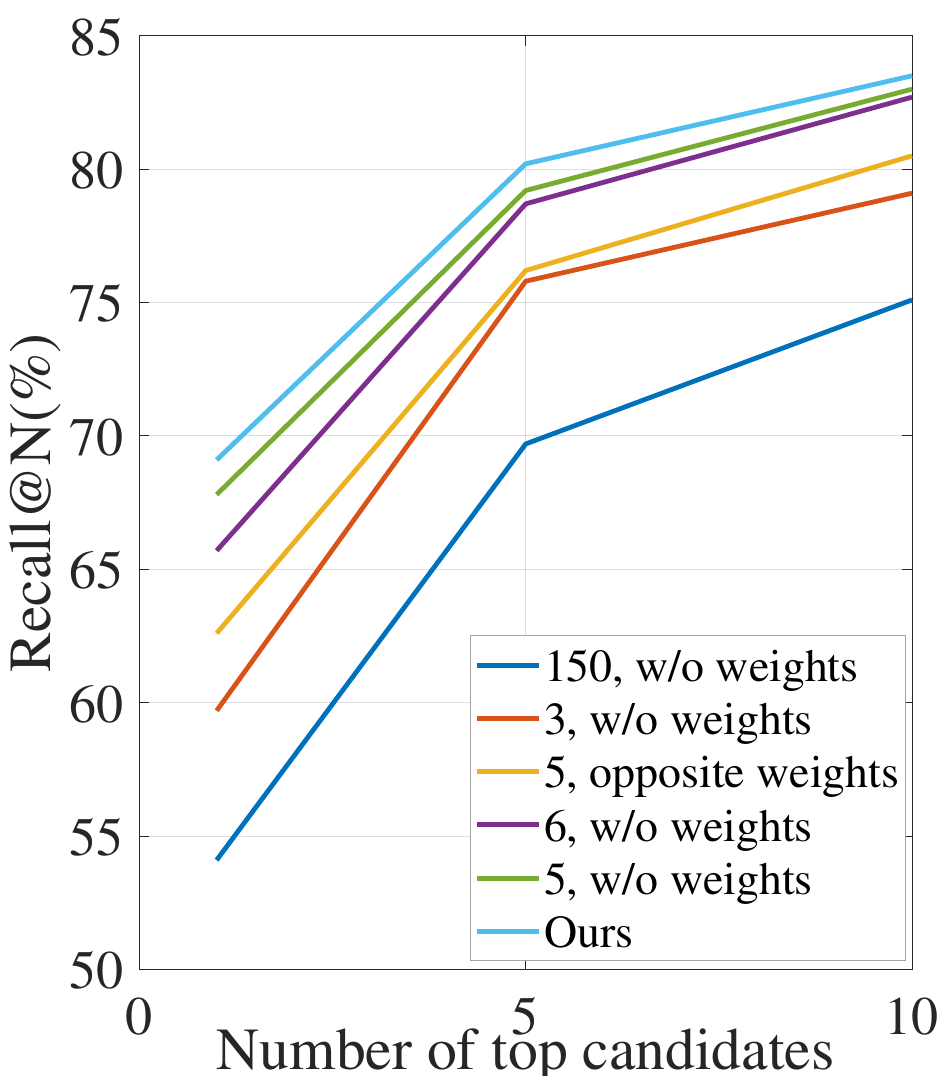}}
    \subfloat[]{
      \label{fig:seg_test}
    \includegraphics[width=0.48\linewidth]{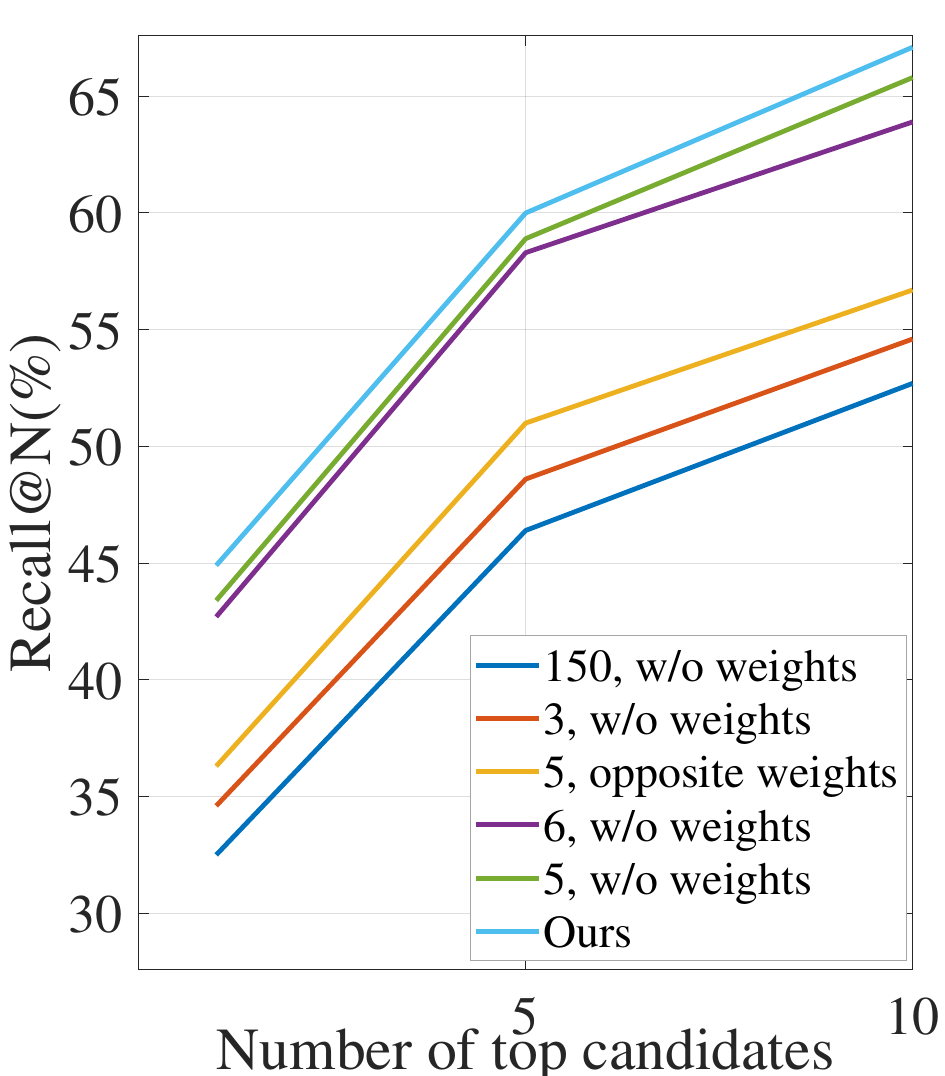}}
      \vspace{-0.2cm}
    \caption{Performance of {seg-branch} using basic global features with different clustered classes on (a) MSLS val. and (b) MSLS test dataset.}
    \label{fig:class}
\end{figure}

\section{Ablation Study of Segmentation Label Map Encoding}


We test $C=3, 6, 150$ in {seg-branch} for the number of clustered classes without weights. After clustering, the redefined 3 categories are \{sky\&ground, dynamic objects, and static objects\}, and the redefined 6 categories are consistent with the classification method in our paper, but dynamic objects are retained.
It can be seen that a too-small value of $C$ will lose the uniqueness of the scene, and the performance is poor when $C=150$, meaning that too fine-grained segmentation makes training difficult.
Comparing the result of $C=5$ and $C=6$, it is better to ignore dynamic objects, which is accorded with our intuition.

\begin{table}[b]
\centering
    \caption{The sample ratio corresponding to different $N_t$}
    \label{stab:group}
  \scalebox{1}{
    \begin{tabular}{c|ccccc}
    \toprule
    $N_t$ &3 & 5 &10 &15 & 20\\ 
    \midrule
      $\mathcal{S}_1$ &60.75\% &64.76\% &69.55\% & 72.11\%&73.82\%\\
      $\mathcal{S}_2$ &39.25\% &35.24\% &30.45\% & 27.89\%&26.18\%\\
    \bottomrule
    \end{tabular}}
\end{table}

In addition, we present the results of different prior weight settings for $C=5$. 
In \fref{fig:class}, the opposite weight configuration means: 0.5 for static and other objects, 1 for sky and ground, and 2 for vegetation.
Comparing the unweighted and weighted cases, we can conclude that using appropriate prior weights can guide model to converge.

\section{Sensitivity to hyper-parameters}
We further evaluate the sensitivity of our model to changes in the other two hyper-parameters: $N_t$ and $N_m$ in our group partition strategy.

\begin{figure}[t]
    \centering
    \includegraphics[width=0.7\linewidth]{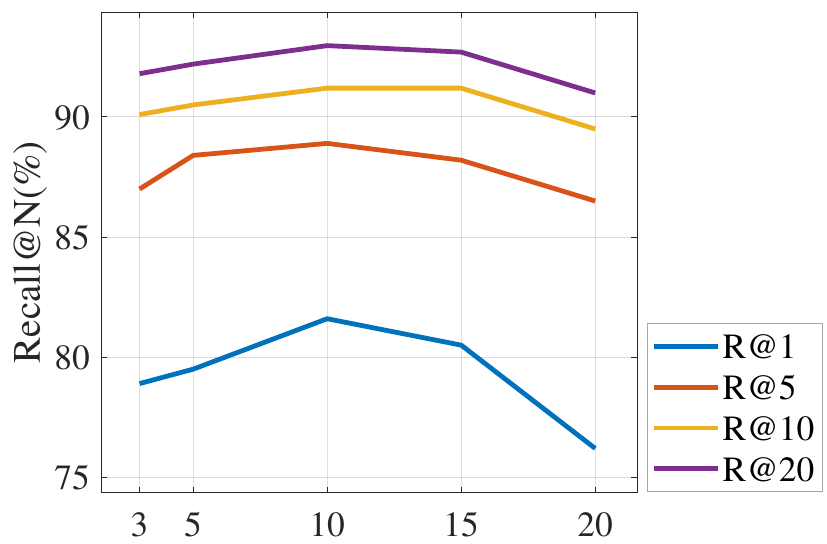}
    \caption{Ablation experiments on the recall performance of StructVPR with different $N_t$.}
    \label{sfig:group}
\end{figure}

Here we perform unweighted selective distillation with the experimental setup of GP-S, that is, on samples belong to $\mathcal{S}_1$.
As shown in \tref{stab:group} and \fref{sfig:group}, the results show the robustness of our method: it can be seen that within the appropriate range, the performance is relatively close. Finally, we select $N_t=10$ in the main paper.
This selection is also consistent with experience, that is, we usually use R@10 as the maximum acceptable metric to evaluate model's performance.

After $N_t$ is set as 10, $N_m$ is mainly used to limit the weight range. 

\begin{figure}[t]
    \centering
    \includegraphics[width=\linewidth]{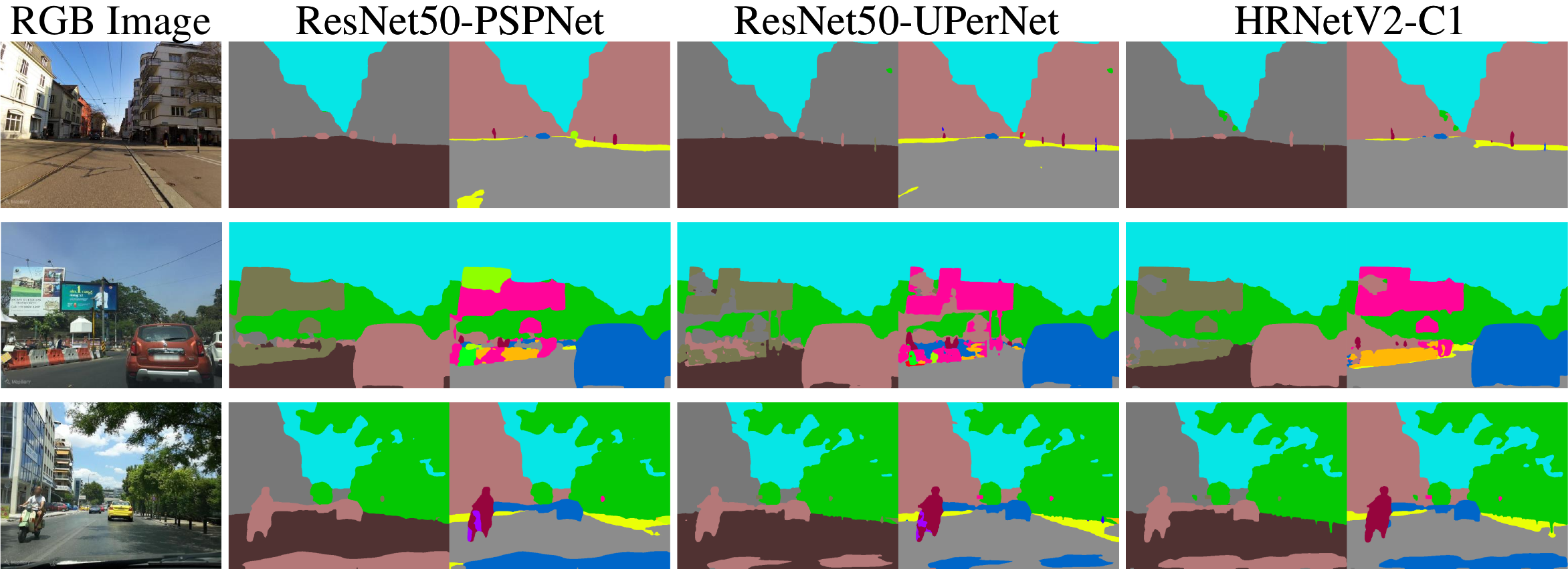}
    \vspace{-0.3cm}
    \caption{Examples of semantic segmentation models with 6/150 classes.}
    \label{sfig:segmodel}
\end{figure}

\section{Ablation Study of  Sensitivity to Segmentation Models}
{In order to use accurate semantic information, some previous works\cite{dasgil,paolicelli2022learning} use synthetic virtual datasets for training, and then generalize to real-world datasets through domain adaptation.}

There are no segmentation ground truths in the VPR datasets, and the segmentation images obtained by the open-source model may not be accurate enough. 
To validate the robustness of StructVPR++ to biased segmentation images, 
we use another two semantic segmentation models, UPerNet and HRNetV2.
We use the same code-base with configuration file of ``ade20k-resnet50-upernet.yaml'' and ``ade20k-hrnetv2.yaml''.
As shown in \fref{sfig:segmodel}, although there are some differences in the 150-class segmentation results (right column) of the three models, while in the case of 6-class segmentation (left column), the difference for the input of seg-branch among these models is smaller.

\begin{table}[b]
  \centering
\caption{Performance of \textit{Seg-branch} and StructVPR++ with Different Segmentation Models. PSP refers to ResNet50-PSPNet (used in the main paper), UPer refers to ResNet50-UPerNet, and HR refers to HRNetV2-C1}
\vspace{-0.3cm}
\label{stab:seg_abl}
  \scalebox{0.8}{
 \renewcommand{\arraystretch}{1.2}
\begin{tabular}{c|ccc|ccc|ccc}
\toprule
\multirow{2}{*}{Method} & \multicolumn{3}{c|}{MSLS val}& \multicolumn{3}{c|}{MSLS test} & \multicolumn{3}{c}{Nordland} \\
\cline{2-10}
&R@1  & R@5  & R@10 &R@1  & R@5  & R@10  &R@1  & R@5  & R@10\\
\midrule
SEG-UPer & 71.6&82.0&84.9&48.3&63.8&\textbf{70.2}&45.2&65.4&73.0\\
SEG-HR & 71.9&81.9&85.0&\textbf{49.0}&\textbf{64.1}&70.0&\textbf{45.6}&65.5&73.4\\
\textbf{SEG-PSP} &\textbf{72.3}&\textbf{82.4}&\textbf{85.3}&48.9&64.0&70.1&45.4&\textbf{65.7}&\textbf{73.5}\\
\midrule
Ours-UPer & 83.5&91.3&{92.9}&65.4&80.8&84.7&58.2&76.6&83.8\\
Ours-HR &82.6 & 90.4&91.9 &\textbf{65.8}& 81.3&85.0 &\textbf{58.7}&\textbf{77.7}&\textbf{84.7} \\
\textbf{Ours-PSP} &\textbf{84.3} & \textbf{91.5} & \textbf{93.1} & {65.7} & \textbf{81.4}& \textbf{85.3}&58.4&77.3&84.1\\
\bottomrule
\end{tabular}}
\end{table}

\begin{figure*}[b]
    \centering
    \includegraphics[width=\linewidth]{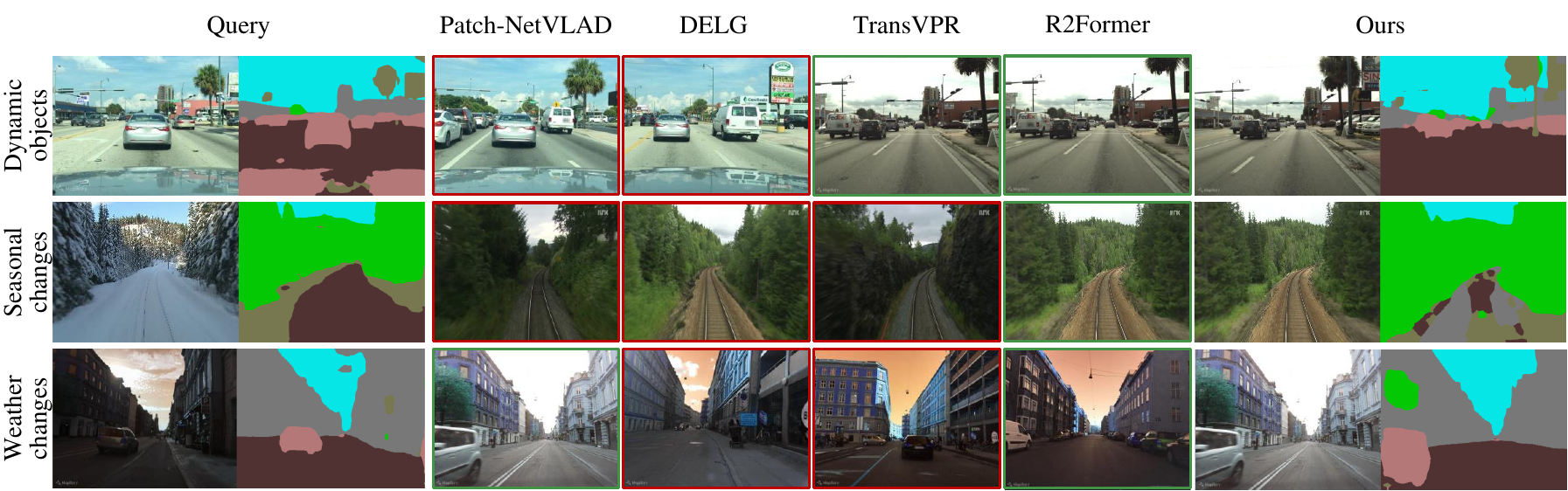}
    \vspace{-0.4cm}
    \caption{In these examples, our method and R2Former both successfully retrieve the matching reference images. For other methods, red borders indicate false matches and green borders indicate correct matches.}
    \label{fig:vis}
\end{figure*}

\begin{figure*}[t]
    \centering
    \includegraphics[width=0.9\linewidth]{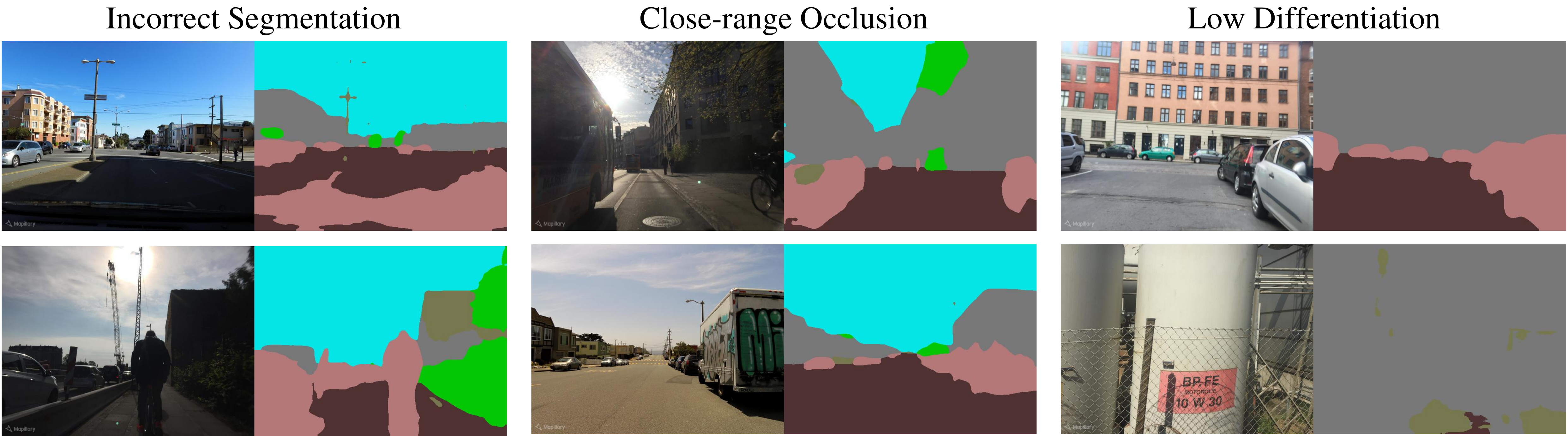}
    \caption{Some typical examples in $\mathcal{D}_4$. This set represents situations where seg-branch performs poorly, including incorrect segmentation, close-range occlusion, and low segmentation differentiation.}
    \label{fig:D4}
\end{figure*}

\begin{figure*}[t]
    \centering
    \includegraphics[width=0.9\linewidth]{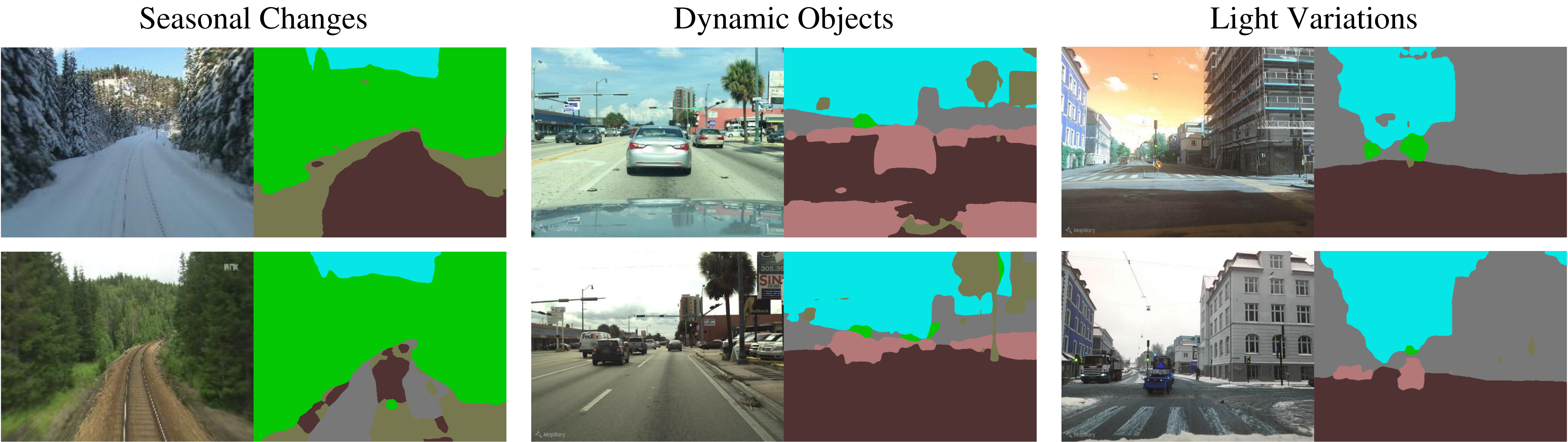}
    \caption{Some typical examples in $\mathcal{D}_1$. This set has a very low percentage of samples, representing situations where seg-branch performs very well but rgb-branch performs poorly.}
    \label{fig:D1}
\end{figure*}

\tref{stab:seg_abl} shows that StructVPR++ is compatible with many models and is little affected by segmentation models.
This also means that StructVPR++ can achieve excellent performance without relying on very precise semantic annotations ground truth.
This is expected since the structural and semantic information we extract does not rely on completely accurate pixel-level segmentation, but more on spatial relative positional relationships within and between images.
Moreover, the clustering operation in SLME also makes StructVPR++ less sensitive to vanilla segmentation results with hundreds of categories.

\section{Additional Qualitative Results}
\label{sec:3}

\fref{fig:vis} illustrates example matches with challenging conditions, such as viewpoint changes, occlusions caused by dynamic objects, and seasons. 
In these examples, other methods show a tendency to retrieve images with similar appearance as the query. Especially in the case of MSLS, the color tone and the vehicle ahead of retrievals of Patch-NetVLAD and DELG are consistent with query.
Ours can successfully retrieve images based on structural information and semantic alignment knowledge, paying more attention to the spatial information of static objects in the background.

For $\mathcal{D}_4$, this set represents situations where seg-branch performs poorly, with no positives retrieved in the top 10, regardless of the performance of the RGB branch. 
Therefore, we give the samples of query in Figure \ref{fig:D4}, showing two examples in each case.
Typical cases include incorrect segmentation from open-source segmentation model, low segmentation differentiation with no semantic features to extract, and close-range occlusion causing segmentation errors.

For $\mathcal{D}_1$, this set has a very low percentage of samples, representing situations where seg-branch performs very well but rgb-branch performs poorly, which mainly occurs when the appearance information of the image changes significantly.
As shown in Figure \ref{fig:D1}, typical cases include seasonal changes, light/weather variations, and dynamic objects that do not affect the segmentation results of building edges.

\bibliographystyle{IEEEtran}
\bibliography{ref}

\end{CJK*}